\documentclass[journal, twoside]{IEEEtran} 

\def\arxivversion{}

\newcommand{\ifarxiv}[2]{\ifdefined\arxivversion#1\else#2\fi}

\usepackage[table]{xcolor}
\usepackage{float}  
\definecolor{wschat}{HTML}{E53935}     
\definecolor{wsedit}{HTML}{FFA656}     
\definecolor{opensource}{HTML}{4CAF50}  
\definecolor{Light1}{rgb}{0.98, 0.95, 0.90}
\definecolor{Light2}{rgb}{0.98, 0.98, 0.93}
\definecolor{Light3}{rgb}{0.98, 0.98, 1}
\definecolor{Light4}{rgb}{0.93, 0.98, 0.98}

\usepackage{graphics} 
\usepackage{epsfig} 
\usepackage{times} 
\usepackage{mathtools}

\usepackage{booktabs}

\usepackage{listings}

\usepackage{algorithm}
\usepackage{algorithmicx}
\usepackage{algpseudocode}

\definecolor{pyblue}{RGB}{0,102,204}      
\definecolor{pygreen}{RGB}{0,128,0}       
\definecolor{pyred}{RGB}{187,0,0}         
\definecolor{pypurple}{RGB}{128,0,128}    
\definecolor{pyorange}{RGB}{255,102,0}    
\definecolor{pygray}{RGB}{128,128,128}    
\definecolor{pybackground}{RGB}{248,248,248} 

\usepackage{makecell}

\usepackage{multicol}

\usepackage[breaklinks,colorlinks]{hyperref}

\usepackage[capitalize]{cleveref}
\Crefname{section}{Sec.}{Secs.}
\Crefname{section}{Section}{Sections}
\Crefname{table}{Table}{Tables}
\Crefname{table}{Tab.}{Tabs.}

\usepackage{threeparttable}

\algnewcommand{\LeftComment}[1]{\Statex \(\triangleright\) #1}

\usepackage{cite}
\usepackage{amsmath,amssymb,amsfonts}
\usepackage{bm} 
\def\BibTeX{{\rm B\kern-.05em{\sc i\kern-.025em b}\kern-.08em
    T\kern-.1667em\lower.7ex\hbox{E}\kern-.125emX}}

\Crefname{equation}{Eq.}{Eqs.} 
\Crefname{figure}{Fig.}{Figs.}

\graphicspath{{}{}{}{}} 

\definecolor{dlrprim1}{HTML}{000000} 
\definecolor{dlrprim2}{HTML}{666666} 
\definecolor{dlrprim3}{HTML}{b9cad2}
\definecolor{dlrprim4}{HTML}{ffffff} 

\definecolor{dlrblue1}{HTML}{00658b} 
\definecolor{dlrblue2}{HTML}{3b98cb}
\definecolor{dlrblue3}{HTML}{6cb9dc}
\definecolor{dlrblue4}{HTML}{a7d3ec}
\definecolor{dlrblue5}{HTML}{d1e8fa}

\definecolor{dlryellow1}{HTML}{d2ae3d}  
\definecolor{dlryellow2}{HTML}{f2cd51} 
\definecolor{dlryellow3}{HTML}{f8de53}
\definecolor{dlryellow3}{HTML}{fcea7a}
\definecolor{dlryellow3}{HTML}{fff8be}

\definecolor{dlrgreen1}{HTML}{82a043} 
\definecolor{dlrgreen2}{HTML}{a6bf51}
\definecolor{dlrgreen3}{HTML}{cad55c}
\definecolor{dlrgreen4}{HTML}{d9df78}
\definecolor{dlrgreen5}{HTML}{e6eaaf}

\definecolor{dlrgray1}{HTML}{666666} 
\definecolor{dlrgray2}{HTML}{868585}
\definecolor{dlrgray3}{HTML}{b1b1b1}
\definecolor{dlrgray4}{HTML}{cfcfcf}
\definecolor{dlrgray5}{HTML}{ebebeb}

\usepackage[textsize=scriptsize]{todonotes} 
\tikzset{
	cross/.pic = {
		\draw[rotate = 45, line width=0.75pt] (-#1,0) -- (#1,0);
		\draw[rotate = 45, line width=0.75pt] (0,-#1) -- (0, #1);
	}
}

\newcommand{\annotategraphics}[5][]{
  \begin{tikzpicture}
    \node[anchor=south west, inner sep=0] (image) at (0,0) {
      #2 
    };
    \begin{scope}[x={(image.south east)},y={(image.north west)}]
      \node[black, text opacity=1, fill=white, fill opacity=0.8, anchor=north west, #1] at (#3, #4) {#5};%
    \end{scope}
  \end{tikzpicture}%
}

\usepackage{orcidlink}

\newcommand{\systemname}{IROSA}

\newcommand{\rmcaffiliation}{German Aerospace Center (DLR),
	Institute of Robotics and Mechatronics (RMC),
	M\"unchener Str. 20, 82234 We\ss ling, Germany. }

\IEEEoverridecommandlockouts                             

\newcommand{\bk}{\bm{k}}
\newcommand{\bK}{\bm{K}}

\newcommand{\mean}{\bm{\mu}}
\newcommand{\covariance}{\bm{\Sigma}}

\newcommand{\inputVariable}{\bm{s}}
\newcommand{\outputVariable}{\bm{\xi}}

\newcommand{\kernelMatrix}{\bK}
\newcommand{\kernelVector}{\bk^*}

\newcommand{\orcidMarkus}{0000-0001-8229-9410}
\newcommand{\orcidEdoardo}{0009-0006-2210-4507}
\newcommand{\orcidAlin}{0000-0001-5343-9074}
\newcommand{\orcidFreek}{0000-0001-9555-9517}
\newcommand{\orcidJoao}{0000-0003-1428-8933}
\newcommand{\orcidSamuel}{0000-0002-7923-8307}
\newcommand{\orcidThomas}{0000-0002-1074-9504}
\newcommand{\orcidGuide}{0009-0002-3521-6517}
\newcommand{\orcidFlo}{0000-0002-5706-3278}
\newcommand{\orcidKorbi}{0000-0002-6016-6235}
\newcommand{\orcidMax}{0000-0002-7635-0248}
\newcommand{\orcidStefan}{0009-0004-5421-9988}
\newcommand{\orcidTimo}{0009-0007-3451-0833}
\newcommand{\addorcidMarkus}{\orcidlink{\orcidMarkus}}
\newcommand{\addorcidAlin}{\orcidlink{\orcidAlin}}
\newcommand{\addorcidFreek}{\orcidlink{\orcidFreek}}
\newcommand{\addorcidJoao}{\orcidlink{\orcidJoao}}
\newcommand{\addorcidSamuel}{\orcidlink{\orcidSamuel}}
\newcommand{\addorcidThomas}{\orcidlink{\orcidThomas}}
\newcommand{\addorcidEdoardo}{\orcidlink{\orcidEdoardo}}
\newcommand{\addorcidGuide}{\orcidlink{\orcidGuide}}
\newcommand{\addorcidFlo}{\orcidlink{\orcidFlo}}
\newcommand{\addorcidKorbi}{\orcidlink{\orcidKorbi}}
\newcommand{\addorcidMax}{\orcidlink{\orcidMax}}
\newcommand{\addorcidStefan}{\orcidlink{\orcidStefan}}
\newcommand{\addorcidTimo}{\orcidlink{\orcidTimo}}

\title{
	MOMO: A framework for seamless physical, verbal, and graphical robot skill learning and adaptation
}

\author{
	Markus Knauer\addorcidMarkus$^{1, 2}$,
	Edoardo Fiorini\addorcidEdoardo$^{1}$,
	Maximilian M\"uhlbauer\addorcidMax$^{2,1}$,
	Stefan Schneyer\addorcidStefan$^{1, 2}$, \\
	Promwat Angsuratanawech\addorcidGuide$^{1, 2}$,
	Florian Samuel Lay\addorcidFlo$^{1}$,
	Timo Bachmann\addorcidTimo$^{1}$,
	Samuel Bustamante\addorcidSamuel$^{1, 2}$, \\
	Korbinian Nottensteiner\addorcidKorbi$^{1}$,
	Freek Stulp\addorcidFreek$^{1}$,
	Alin Albu-Sch\"affer\addorcidAlin$^{1, 2}$,
	Jo\~ao Silv\'erio\addorcidJoao$^{1}$,
	and Thomas Eiband\addorcidThomas$^{1}$
		\thanks{This work was partially funded by the DLR project ``ASPIRO''; the European Union's Horizon Research and Innovation Program under Grant 101136067 (INVERSE); the Federal Ministry for Economic Affairs and Climate Protection with DARP funds based on a decision by the German Bundestag and by the European Union - NextGenerationEU; and partially supported by the German Federal Ministry of Research, Technology and Space (BMFTR) under the Robotics Institute Germany (RIG).}
		\thanks{$^{1}$ All authors are with the \rmcaffiliation \texttt{\{first\}.\{last\}@dlr.de}}
		\thanks{$^{2}$ Markus Knauer, Maximilian M\"uhlbauer, Stefan Schneyer, Promwat Angsuratanawech, Samuel Bustamante and Alin Albu-Sch\"affer are also with the School of Computation, Information and Technology (CIT), Technical University of Munich (TUM), Arcisstr. 21, 80333 Munich, Germany. \texttt{m.knauer@tum.de}}
		\ifarxiv{\thanks{Digital Object Identifier (DOI): 10.1109/RAP.2026.3724466}}{}
}

\begin{document}

\maketitle

\begin{abstract}
	\label{sec:abstract}
	Industrial robot applications require increasingly flexible systems that non-expert users can easily adapt for varying tasks and environments. However, different adaptations benefit from different interaction modalities.
	We present an interactive framework that enables robot skill adaptation through three complementary modalities: kinesthetic touch for precise spatial corrections, natural language for high-level semantic modifications, and a graphical web interface for visualizing geometric relations and trajectories, inspecting and adjusting parameters, and editing via-points by drag-and-drop. The framework integrates five components: energy-based human-intention detection, a tool-based LLM architecture (where the LLM selects and parameterizes predefined functions rather than generating code) for safe natural language adaptation, Kernelized Movement Primitives (KMPs) for motion encoding,
	probabilistic Virtual Fixtures for guided demonstration recording, and ergodic control for surface finishing. We demonstrate that this tool-based LLM architecture generalizes skill adaptation from KMPs to ergodic control, enabling voice-commanded surface finishing. Validation on a 7-DoF torque-controlled robot at the Automatica 2025 trade fair demonstrates the practical applicability of our approach in industrial settings.
\end{abstract}


\section{Introduction}
\label{sec:introduction}

Modern industrial robot applications require flexible systems that can be easily adapted for varying tasks and environments, requiring frequent skill adjustments in response to changing product specifications, environmental conditions, or operational requirements~\cite{Villani2018}. However, adapting robot skills often requires manual reprogramming, a complex and time-consuming process accessible only to expert programmers~\cite{Billard2008, Ravichandar2020}. This creates a bottleneck: non-expert operators lack intuitive tools to quickly modify robot behaviors, resulting in extended downtime and reduced flexibility. Making robot skill adaptation accessible to non-experts is therefore essential for deploying adaptive robotic systems in industrial settings.
\begin{figure}[ht]
	\centering
	\includegraphics[width=\columnwidth]{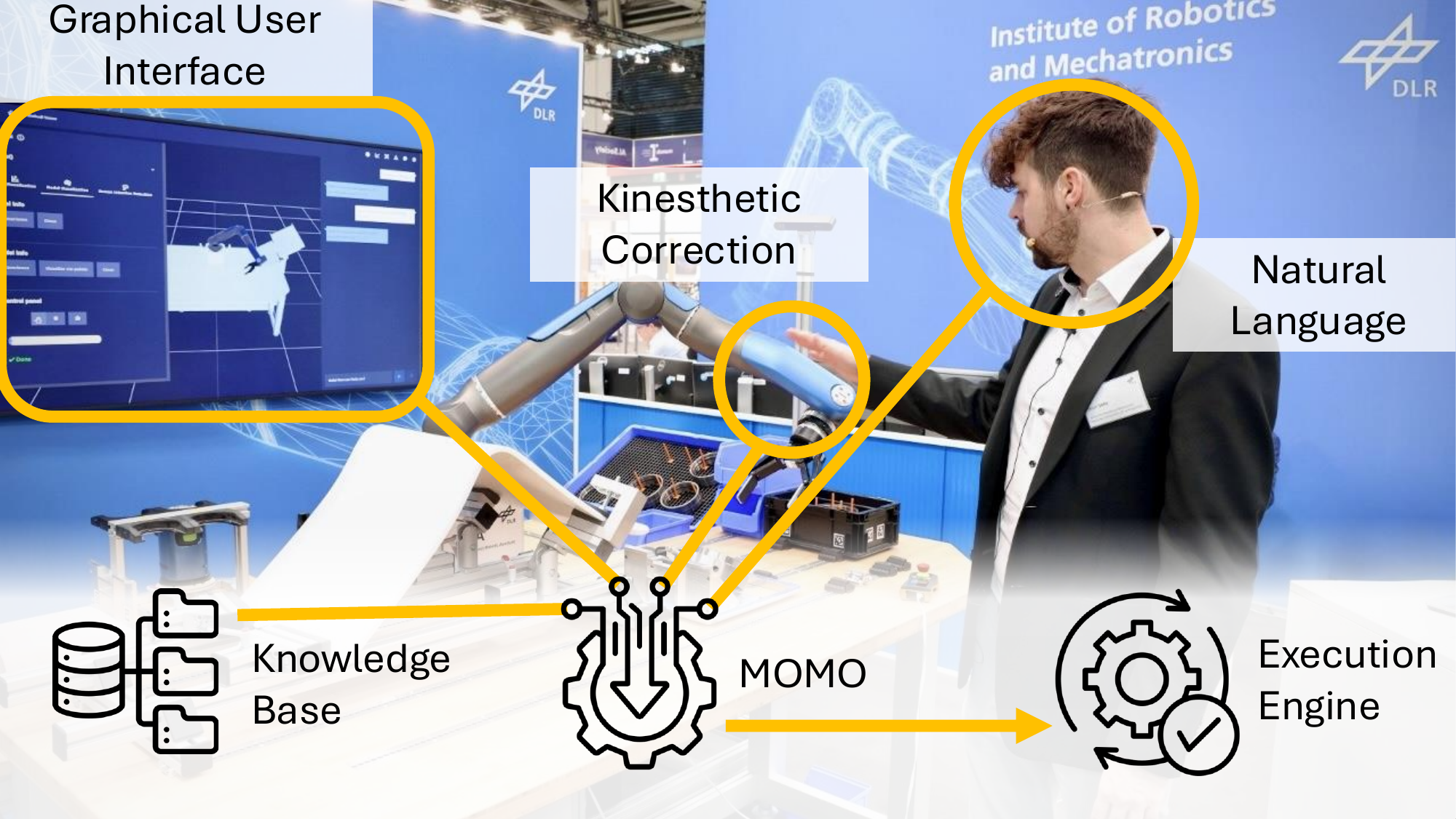}
	\caption{\label{fig:graphical_abstract} Overview of the framework. The three interaction modalities (physical, verbal, and graphical) enable complementary skill adaptation, demonstrated on industrial robot tasks at a trade fair. The \textbf{MOMO} (\textbf{Mo}tion \textbf{Mo}dulation) module manages the different inputs and adapts the skill accordingly. The Execution Engine deploys the resulting trajectories on the robot.}
\end{figure}

Recent advances in Large Language Models (LLMs) show potential for more intuitive robot skill adaptation through natural language interaction, allowing users to communicate desired modifications in everyday language rather than specialized programming syntax. Unlike traditional teach-pendant programming, end-users can express their intentions conversationally, with the system interpreting and translating these into appropriate robot actions.
However, LLMs are not optimal for all adaptation scenarios. While they excel at high-level semantic concepts, they may be unnecessarily indirect for fine-tuning or simple adjustments. Different adaptation tasks thus benefit from different interaction modalities: kinesthetic correction provides natural spatial feedback, graphical interfaces enable parameter inspection and adjustment, and natural language allows for intuitive semantic-level modifications. Rather than adopting a one-size-fits-all approach, the choice of interaction method should rest with the user.
Furthermore, users have different strengths in using these interaction methods. For instance, a user might feel uncomfortable with physically correcting a moving robot and may prefer natural language interaction instead.

We propose an interactive framework enabling robot skill adaptation through multiple complementary modalities: kinesthetic feedback for force-guided corrections, natural language for high-level semantic modifications, and graphical interfaces for parameter adjustment and visualization. Fine-tuning and small geometric adjustments are naturally performed via kinesthetic correction, larger task-level adaptations (obstacle avoidance, trajectory modifications) are efficiently expressed through natural language, and the graphical interface enables structured parameter inspection and adjustment.
The main contributions of our work are:
\begin{enumerate}
	\item A unified multi-modal robot skill adaptation framework combining physical interaction, natural language (voice and text), and a web-based graphical interface with drag-and-drop via-point editing, enabling end-users to select the modality best suited to each adaptation task and their personal strengths;
	\item Integration of five complementary components: (i) energy-based human intention detection for kinesthetic corrections~\cite{Fiorini2026}, (ii) a tool-based LLM architecture for safe natural language adaptation~\cite{KnauerIROSA2026}, (iii) KMP-based skill representation with online via-point adaptation~\cite{Huang2019}, (iv) probabilistic Virtual Fixtures for guided demonstration recording~\cite{muehlbauer2024probabilistic,muehlbauer2025unified}, and (v) ergodic control for surface finishing~\cite{Schneyer2025}, demonstrating that the tool-based LLM architecture generalizes across skill representations;
	\item An open-source release of the \systemname{} and KMP/TP-KMP with runnable examples (Sec.~S-I).
\end{enumerate}
We validate the framework on a 7-DoF torque-controlled robot performing industrial manipulation tasks. By combining established algorithms with deployment on real industrial hardware, we provide an overview of current trends in intuitive programming of cobots, using our Automatica 2025 trade fair demonstrator as an integrated case study, and demonstrate that interactive multi-modal robot adaptation is viable in production environments and accessible to non-expert users.

\section{Related Work}
\label{sec:related_work}

Multi-modal interfaces for robot programming have gained attention in making skill acquisition more accessible. Early work focused on single modalities: kinesthetic teaching~\cite{Billard2008}, natural language commands~\cite{Tellex2011}, or graphical interfaces~\cite{Alexandrova2015}. Advances in Learning from Demonstration~\cite{Ravichandar2020} have enabled flexible skill representations that can be adapted online.
Large Language Models have been applied to robot control through code generation~\cite{Liang2023} and tool-augmented approaches~\cite{Schick2023, Wang2024}. OVITA~\cite{Maurya2025} uses LLMs for open-vocabulary trajectory adaptation via code synthesis, trading safety for flexibility.
Kernelized Movement Primitives~\cite{Huang2019} provide a probabilistic, non-parametric framework for encoding demonstrations that naturally supports via-point-based adaptation. Physical human-robot interaction for skill correction has been explored using energy-based methods~\cite{Khoramshahi2020}, enabling robots to distinguish intentional corrections from task-related forces. Virtual Fixtures~\cite{Rosenberg1993} provide haptic guidance during teleoperation and demonstration recording, reducing operator workload and improving demonstration consistency.
Ergodic control~\cite{Mathew2011, Ivic2017} offers an alternative to movement primitives for coverage tasks, enabling online adaptation to perturbations. The Spectral Multiscale Coverage (SMC) approach~\cite{Mathew2011} achieves ergodic coverage by minimizing a cost function that compares the Fourier coefficients of the target distribution with those of the observed coverage, generating control actions online at each time step. Despite these advances, no existing framework unifies physical, verbal, and graphical interaction modalities for both skill acquisition and adaptation.


\section{Methodology}
\label{sec:methodology}

We present a unified framework that enables robot skill acquisition and adaptation through three interaction modalities: physical interaction, natural language, and graphical interface. Figure~\ref{fig:graphical_abstract} shows the system architecture and the five integrated components. At the core, the \textbf{MOMO} (\textbf{Mo}tion \textbf{Mo}dulation) module manages the different inputs and adapts the skill accordingly. The Execution Engine then deploys the resulting trajectories on the torque-controlled robot. For motion encoding, Kernelized Movement Primitives (KMPs)~\cite{Huang2019} encode demonstrations as a probabilistic mapping from input $\inputVariable$ to output $\outputVariable$. Given reference data $\mathcal{D} = \{\inputVariable_n, \mean_n, \covariance_n\}_{n=1}^N$, with means $\mean_n$ and covariances $\covariance_n$ extracted via Gaussian Mixture Models from demonstrations, the predictive mean evaluates to:
\begin{equation}
	\mathbb{E}[\outputVariable(\inputVariable^*)] = \kernelVector (\kernelMatrix + \lambda_1 \covariance)^{-1} \mean
\end{equation}
where $\lambda_1$ is a small regularization factor, $\kernelMatrix$ and $\kernelVector$ are kernel matrices measuring distances between training points and between training and test points, respectively (see~\cite{Huang2019} for details).
Via-points $\bar{\mathcal{D}} = \{\bar{\inputVariable}, \bar{\mean}, \bar{\covariance}\}$ can be added online after learning to adapt the trajectory, with small $\bar{\covariance}$ enforcing precise passage through $\bar{\mean}$. This framework has been extended with interactive incremental learning and local trajectory modulation for generalizable skill adaptation~\cite{Knauer2025}. For surface finishing, we extended SMC~\cite{Schneyer2025} by incorporating a variable tool contact area that adapts to nontrivial surface geometries, and learning both desired coverage distributions and preferred motion directions from human demonstrations.

\subsection{Skill Acquisition with Virtual Fixtures}
A Virtual Fixtures component~\cite{muehlbauer2024probabilistic,muehlbauer2025unified} combines dynamical system-based, position-based, and visual servoing fixtures. These fixtures provide probabilistic wrenches fused via Product of Experts arbitration. Variable impedance control~\cite{muehlbauer2025unified} with coupled stiffness matrices guides users during demonstration recording, reducing workload while maintaining precision, with closed-loop passivity ensured by~\cite{muehlbauer2026stabilizing}.
During demonstration recording, these probabilistic Virtual Fixtures assist the operator with haptic guidance, for instance while picking or placing objects.
We use position-based trajectory and dynamical system-based velocity fixtures learned from demonstrations with the variable impedance formulation~\cite{muehlbauer2025unified,muehlbauer2026stabilizing}.
The resulting fixtures, as shown in Fig.~S4 in the supplementary material, support the operator in precise alignment, thus reducing operator workload and improving demonstration consistency, particularly for industrial tasks requiring high precision.
Subsequent corrections of the encoded motion are still possible through physical interaction with the impedance controller during execution.

\subsection{Physical Interaction Modality}
For intuitive fine-tuning during execution, users can physically interact with the robot to correct trajectories. We employ an energy-tank-based intention detection~\cite{Fiorini2026}, where an energy tank is assigned to each degree of freedom (DoF) to track interaction dynamics, mapping the user-injected energy to an intention index in the range $[0,1]$. The tank's dissipation is adapted by the KMP uncertainty. High-variance regions dissipate energy faster, yielding shorter compliant windows for corrections, while low-variance regions dissipate more slowly, giving the operator more time for fine, precise corrections.
The robot achieves compliance and facilitates physical contact in the relevant DoFs through varying stiffness based on the intention index (Fig.~S9). Once the index exceeds a predefined threshold, DoF-specific via-points are automatically inserted into the underlying KMP. This enables precise, local corrections without affecting distant trajectory segments.

\subsection{Natural Language Modality}
Users adapt skills through voice or text commands processed by a tool-based LLM architecture~\cite{KnauerIROSA2026}, where the LLM selects and parameterizes predefined functions rather than generating executable code. This maintains strict separation between language understanding and robot control. The \systemname{} is accessible via an integrated chat window (Fig.~S3 in the supplementary material) and supports skill selection and adaptation through natural language. The LLM selects appropriate tools and extracts parameters from the instruction and environmental context. Each tool is pre-validated by the developer with defined parameter bounds and safety constraints, ensuring that only verified operations reach the robot.

\paragraph*{Adapt KMP-Based Skills}
For KMP-based skill adaptation, the available tools include speed modulation (e.g., ``Go slower near the box''), trajectory correction via via-point insertion (e.g., ``Move more to the left at the start''), and obstacle avoidance using signed distance field repulsion (e.g., ``Avoid the blue box'').

\paragraph*{Parameterize Ergodic Surface Finishing}
We extend the tool-based LLM architecture to ergodic surface finishing~\cite{Schneyer2025},
which provides a framework for generating trajectories that distribute coverage intensity
according to task-specific spatial objectives. The approach learns desired coverage distributions and motion preferences from demonstrations and incorporates tool imprint models for accurate coverage on complex geometries.
Because trajectories are generated online, the approach naturally supports human intervention and parameter adaptation.
We define tools for start/stop execution, tool velocity, impedance stiffness, and contact force control.

This demonstrates that the tool-based architecture is not tied to a specific skill representation, as shown in \Cref{fig:slow_down_experiment} and Fig.~S13 in the supplementary material, where the same chat interface controls a surface finishing task instead of a KMP-based motion. The same LLM interprets user intent and selects appropriate tools, while the underlying control method determines how those tools affect the robot behavior.

\subsection{Graphical Interface Modality}
A web-based user interface provides visualization, parameter inspection, and adjustment for different skill representations.
It features the visualization of the workcell and a virtual robot arm. Additionally, an interaction panel is organized into four functional tabs: \textit{Default Fixtures}, \textit{Data Visualization}, \textit{Model Visualization}, and \textit{Human Intention Detection}.
For more general information about our Human Factory Interface (HFI), see Sec.~S-II in the supplementary material.
The Data Visualization tab (Fig.~S5 in the supplementary material) allows inspection of recorded demonstrations, displaying the Cartesian trajectory in the virtual environment. Users can select individual demonstrations and switch between visualization modes including Cartesian position, Cartesian pose with position and orientation, force profile, and joint configuration. The joint configuration view (Fig.~S6) renders a robot digital twin at the selected time step, enabling intuitive inspection of the robot pose throughout the demonstration.
The Model Visualization tab (Fig.~S7) displays the learned KMP mean and covariance, providing a direct view of the skill's underlying data representation. Users can compare the original and updated model side by side, e.g., to verify the effect of via-point insertions. The tab includes an edit mode for interactive trajectory modification and a trajectory control panel for starting execution and monitoring progress. During execution (Fig.~S8), a digital twin of the robot moves in sync with the real robot, and the progress bar indicates the current execution state.
Users add via-points by drag-and-drop on the trajectory (Fig.~S12), and adapt or delete existing ones via a right-click menu (Fig.~S11); the adapted trajectory (yellow) is shown against the original (blue) in Fig.~S10. The interface provides visual feedback that is difficult to convey through physical or verbal interaction: users can see exactly where via-points will affect the trajectory and verify that the adapted path meets requirements before execution.

\subsection{Complementary Modalities}
Each modality offers distinct advantages:
\begin{itemize}
	\item \textbf{Physical interaction} excels at precise spatial corrections.
	Users directly see and feel the robot's motion and apply corrections intuitively. However, it requires physical proximity and good physical guidance skills, and is limited to local adjustments.
	\item \textbf{Natural language} efficiently expresses high-level semantic modifications (``avoid the obstacle'') and references to perceptual features (``move closer to the red box''). It is less suited for precise geometric specifications.
	\item \textbf{Graphical interface} enables methodical parameter adjustment and provides visual confirmation of planned trajectories. Users can see all via-points simultaneously and adjust them with a precision that is difficult to achieve through verbal description or physical interaction.
\end{itemize}
Users freely switch between modalities based on the task at hand and personal skills. An operator might first use voice commands to add obstacle avoidance, then fine-tune the trajectory via kinesthetic correction, and finally verify changes through the graphical interface.

\section{Evaluation}
\label{sec:evaluation}

We validate our framework through live demonstration at the Automatica 2025 trade fair, where visitors could interact with the system using all three modalities. A video of the demonstration is provided in the supplementary materials.

\subsection{Experimental Setup}
\begin{figure}
	\centering
	\annotategraphics{\includegraphics[width=0.38\columnwidth]{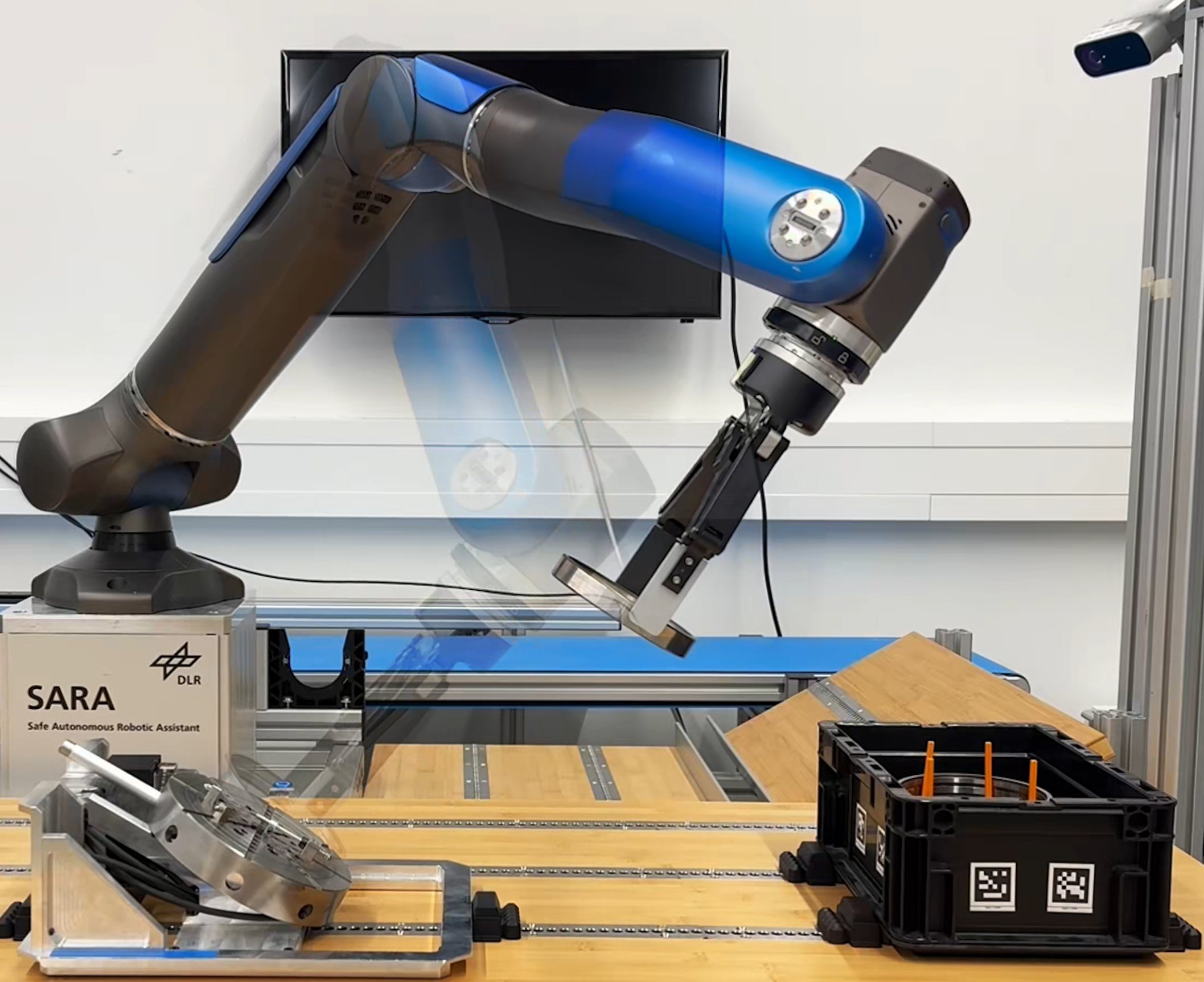}}{0.05}{0.95}{(a)}%
	\hfill%
	\annotategraphics{\includegraphics[width=0.59\columnwidth]{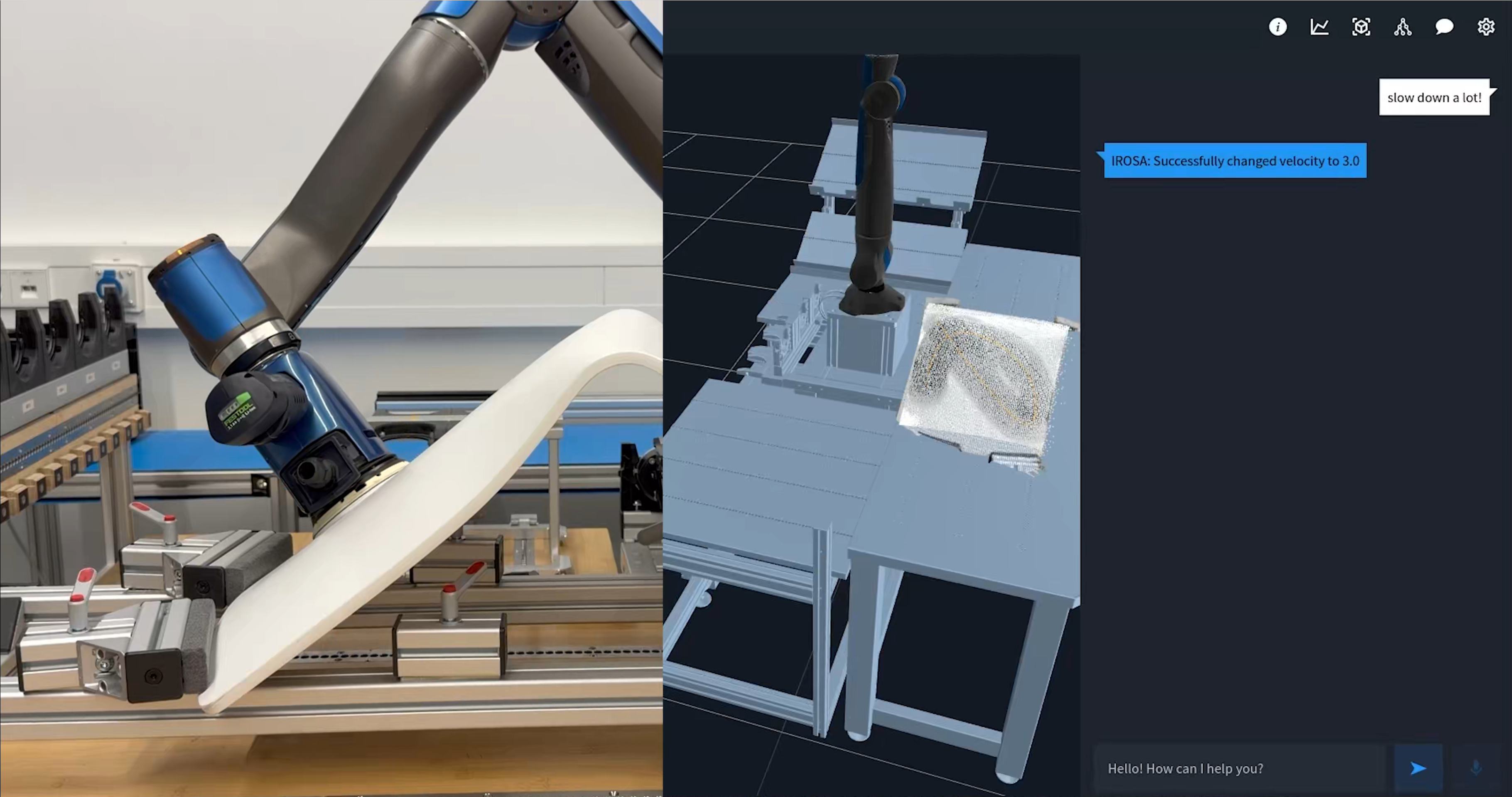}}{0.03}{0.95}{(b)}
	\caption{(a) Bearing ring insertion task: the transparent robot shows original speed, the opaque one shows the adapted trajectory after the voice command ``slow down between box and station.'' (b) Ergodic surface finishing task with the LLM chat interface issuing a similar slow-down command, alongside a virtual twin of the robot workcell.}
	\label{fig:slow_down_experiment}
\end{figure}

The framework operates a 7-DoF torque-controlled robot shown in \Cref{fig:graphical_abstract}, performing industrial tasks including a bearing ring insertion and surface finishing (\Cref{fig:slow_down_experiment}). The system runs on industrial hardware
with a local LLM backend (Qwen2.5-VL-72B-Instruct) for natural language processing, ensuring data privacy and low latency. Screenshots of the web-based graphical interface are provided in Sec.~S-III in the supplementary material.

\subsection{Operational Workflow}
The complete workflow consists of the following phases.

\paragraph{Skill Acquisition} An operator records demonstrations via kinesthetic teaching, assisted by probabilistic Virtual Fixtures~\cite{muehlbauer2025unified} that provide haptic guidance. The fixtures combine dynamical system-based guidance for coarse trajectory following with position-based constraints for precise waypoint passage, reducing teaching effort while ensuring consistent demonstration quality.

\paragraph{Skill Execution} The learned KMP generates smooth trajectories adapted to the current object locations.
For surface finishing, ergodic control~\cite{Schneyer2025} ensures uniform coverage while respecting learned motion preferences. Both representations support online adaptation during execution.

\paragraph{Skill Adaptation} When task requirements change (new obstacles, shifted targets, different surface coverage requirements), users adapt the skill through one or more of their preferred modalities, with iterative cycles of execution and adaptation.

\subsection{Qualitative Observations}
During the trade fair demonstration, we observed clear patterns in modality preferences among visitors:
\begin{itemize}
	\item \textbf{Fine-tuning}: Users intuitively used physical interaction for small spatial adjustments, benefiting from direct proprioceptive feedback.
	\item \textbf{Semantic modifications}: Voice commands were preferred for high-level changes such as ``avoid the red box'' or ``move higher at the start'', which are cumbersome to express through physical interaction.
	\item \textbf{Parameter inspection}: The graphical interface was frequently used to visualize trajectories before and after modifications (Fig.~S10 in the supplementary material), and to systematically adjust via-point positions through drag-and-drop (Fig.~S12).
\end{itemize}

\section{Discussion}
\label{sec:discussion}

The framework demonstrates that multi-modal interaction makes robot skill adaptation accessible to non-experts, as demonstrated with trade fair visitors.
Rather than forcing users to adopt a single interaction paradigm, the system allows selection of the method best suited to each task and user-specific skills: physical interaction for precise local corrections, natural language for semantic-level modifications, and the graphical interface for systematic exploration and verification.

A key design decision is integrating multiple approaches that complement each other throughout the skill lifecycle: Virtual Fixtures~\cite{muehlbauer2025unified} assist during demonstration recording, KMPs~\cite{Huang2019} provide a flexible skill representation supporting via-point adaptation from all modalities, and the tool-based LLM architecture~\cite{KnauerIROSA2026} ensures safe natural language processing by constraining the LLM to predefined functions. While the current tool set is bounded, using an LLM rather than conventional keyword-based voice recognition~\cite{Zinchenko2017} provides tolerance to imprecise or varied phrasing, natural context handling, and easy extension, as new tools can be added without retraining a domain-specific speech model.

Extending the LLM tool architecture to ergodic control applications~\cite{Schneyer2025} validates generalizability (Fig.~S13 in the supplementary material). The same natural language interface that adapts KMP-based pick-and-place skills (via-point insertion, obstacle avoidance, speed modulation) also controls surface finishing parameters (coverage velocity, contact force, impedance stiffness). This suggests that the tool-based architecture can accommodate diverse skill representations beyond movement primitives.

The natural language interface is currently limited to predefined tool functions; truly open-ended adaptations require code generation approaches that sacrifice safety guarantees.
Furthermore, physical interaction modalities require force-torque sensing hardware. An end-effector sensor suffices but limits interaction to the last link, whereas joint torque sensors enable whole-body contact detection.
Lastly, Virtual Fixtures assist but do not guarantee demonstration quality, and poor demonstrations still propagate to learned skills.

\section{Conclusion}
\label{sec:conclusion}

We presented a framework for interactive robot skill acquisition and adaptation, combining three interaction modalities: physical interaction for precise spatial corrections, natural language for high-level semantic modifications, and a web-based graphical interface for parameter inspection and adjustment. The framework integrates five complementary components that support each other throughout the skill lifecycle. Live demonstration at Automatica 2025 on a 7-DoF torque-controlled robot validated practical applicability, allowing visitors to naturally switch between modalities based on their needs. Future work will investigate automatic modality suggestion based on task characteristics and conduct a formal user study to quantify modality preferences.



\bibliographystyle{IEEEtran}
\bibliography{bibliography}

\clearpage
\onecolumn

{\centering
  {\Large\textbf{Supplementary Material}\par}
  \vspace{0.5em}
  {\normalsize A framework for seamless physical, verbal, and graphical robot skill learning and adaptation\par}
}
\vspace{1.5em}

\setcounter{section}{0}
\setcounter{figure}{0}
\setcounter{table}{0}
\renewcommand{\thefigure}{S\arabic{figure}}
\renewcommand{\thesection}{S-\Roman{section}}
\renewcommand{\thetable}{S-\Roman{table}}

\newlength{\overviewfigwidth}
\setlength{\overviewfigwidth}{0.9\textwidth}
\newlength{\screenshotfigwidth}
\setlength{\screenshotfigwidth}{0.9\textwidth}

\section{Implementation Details}
\label{sec:implementation}

\begin{multicols}{2}
\subsection{Software Architecture and Technology Stack}
\begin{figure}[H]
	\centering
	\includegraphics[width=\columnwidth]{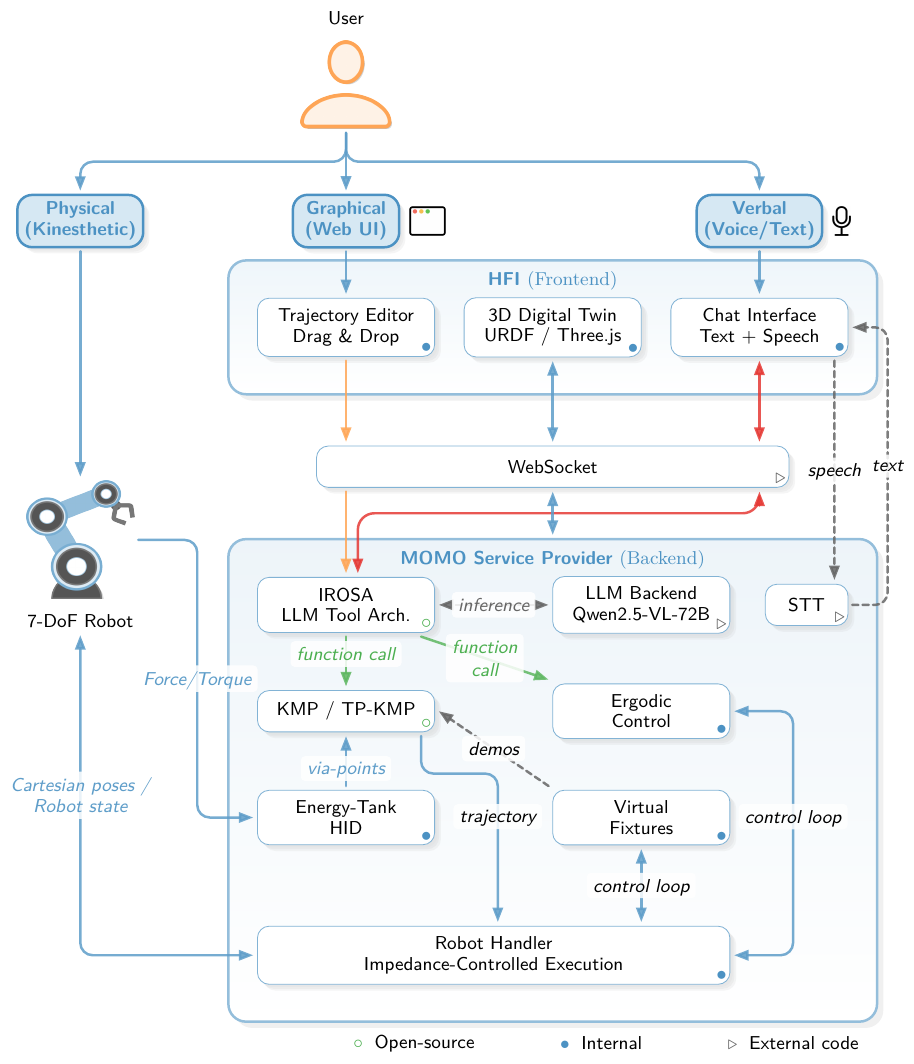}
	\caption{Software architecture and data flow. The three user modalities (physical, verbal, graphical) connect to the system through distinct paths. The HFI Frontend communicates with the MOMO backend via a WebSocket bridge: the \textcolor{wsedit}{orange path} carries trajectory adaptations from the Trajectory Editor, the \textcolor{wschat}{red path} carries bidirectional chat messages to and from the \systemname{}. The Chat Interface accepts text input directly or forwards speech audio to the STT module for transcription. The \systemname{} uses an external LLM backend for inference and dispatches \textcolor{opensource}{function calls} to the skill modules (KMP/TP-KMP, Ergodic Control). The Energy-Tank Human Intention Detection translates Force/Torque readings from the robot into via-points for KMP, while Virtual Fixtures provide demonstration data to KMP and maintain a closed control loop with the Robot Handler. KMP generates trajectories for the Robot Handler, which performs impedance-controlled execution. The Ergodic Control module likewise maintains a closed control loop with the Robot Handler. The Robot Handler sends Cartesian poses to the robot, which in turn provides its state back to the Robot Handler. Component ownership is indicated by symbols: $\circ$~open-source, $\bullet$~internal, $\triangleright$~external code.}
	\label{fig:system_architecture}
\end{figure}

\Cref{fig:system_architecture} shows the software architecture and data flow. The system is built on the following technology stack:

\textbf{Middleware:} All components communicate via \textit{links and nodes} (LN)~\cite{LinksAndNodes2020}, an open-source\footnote{\label{fn:ln}\url{https://gitlab.com/links_and_nodes/links_and_nodes}} (GPLv3) real-time-capable middleware for distributed robotic systems. LN is both a process manager with a graphical user interface and a communication framework providing publish-subscribe (via topics) and request-response (via services) communication patterns. A WebSocket bridge translates the LN protocol to JSON-over-WebSocket, enabling browser-based frontends.

\textbf{MOMO Service Provider:} The central orchestrator registers 16 services over LN: 13~motion modulation services (trajectory data retrieval, model visualization, trajectory execution control, via-point management from all three modalities, and human intention detection) and 3~LLM services (query input, answer retrieval, speech-to-text). It publishes real-time status on two topics: trajectory execution progress and LLM notification events.

\textbf{LLM Backend:} Qwen2.5-VL-72B-Instruct is deployed locally via vLLM on GPU hardware. It was already available on our on-premise inference infrastructure and, among the models we tried~\cite{KnauerIROSA2026}, it selected tools most reliably while supporting OpenAI-compatible function calling, on which the tool-based architecture depends. Its vision-language capability is not exercised in the pipeline reported here, as object positions and semantic labels are passed to the LLM as text, not as images; a vision-language model was nonetheless retained so that visual grounding can be added later without changing the tool interface. The LLM receives tool definitions as typed function schemas (compatible with the OpenAI function calling API) and returns structured tool calls with validated parameters. An example script for running the LLM backend against the \systemname{} is included in the open-source release (\Cref{tab:opensource}).

\textbf{STT:} Speech-to-text transcription via OpenAI Whisper, deployed as a FastAPI service.

\textbf{Robot Handler} (corresponding to the Execution Engine in the graphical abstract of the main paper)\textbf{:} Impedance-controlled~\cite{hogan1984impedance} trajectory execution on a 7-DoF torque-controlled robot~\cite{Iskandar2023}, with force/torque sensing via a momentum-based external observer for physical interaction detection~\cite{Iskandar2021}.

\subsection{IROSA: LLM Tool Architecture}
The IROSA (Interactive Robot Skill Adaptation) architecture defines robot skill adaptations as typed tool functions that the LLM selects and parameterizes. Each tool specifies: a function name, a natural language description, typed parameters with JSON schema, and validated bounds enforced before execution. \Cref{tab:llm_tools} lists all tools.
\vspace{1em}

The LLM receives the tool definitions via the standard function calling API. Given a user command, it returns a structured \texttt{tool\_call} containing the selected function name and extracted arguments. The MOMO module validates all arguments against the defined bounds before executing the tool on the KMP or ergodic model. This separation ensures that the LLM never directly controls the robot---it can only select from pre-validated operations.

New tools can be added by defining a Python function with parameter validation and registering it with a JSON schema description. No LLM retraining is required.

The bounds in \Cref{tab:llm_tools} are fixed constants that do not vary per task, and they were determined empirically. For the ergodic tools we started from the parameter values used in the demonstrations (velocity~6, contact force~15\,N, stiffness~1000\,N/m) and bracketed each with a conservative margin around the range that produced sensible behavior during development; the KMP repulsion radius and speed modulation range were fixed the same way. The bounds are therefore tied to this cell, tooling, and skill set; none of them follow from first principles. Transferring the framework to another setup means re-fitting them once per tool, a configuration change that affects neither the LLM nor its prompts.

\begin{table}[H]
	\centering
	\caption{LLM tool specifications with validated parameter bounds.}
	\label{tab:llm_tools}
	\footnotesize
	\setlength{\tabcolsep}{3pt}
	\begin{tabular}{@{}llll@{}}
		\toprule
		\textbf{Tool} & \textbf{Parameters} & \textbf{Range} & \textbf{Skill} \\
		\midrule
		\rowcolor{Light1}
		AddViaPoints & time $\in [0,1]$, pos.\ $[x,y,z]$ & -- & KMP \\
		AddRepulsion & pos.\ $[x,y,z]$, radius & $r \leq 1.0$\,m & KMP \\
		\rowcolor{Light1}
		SlowDown & percentage, $t_\text{start}$, $t_\text{end}$ & $[1,100]\%$ & KMP \\
		SpeedUp & percentage, $t_\text{start}$, $t_\text{end}$ & $[1,100]\%$ & KMP \\
		\midrule
		\rowcolor{Light1}
		SetVelocity & velocity & $[3, 16]$ & Ergodic \\
		SetForce & force (N) & $[5, 30]$ & Ergodic \\
		\rowcolor{Light1}
		SetStiffness & stiffness (N/m) & $[500, 2000]$ & Ergodic \\
		SetExecState & ``pause'' / ``resume'' & -- & Ergodic \\
		\bottomrule
	\end{tabular}
\end{table}

\subsection{Reliability, Failure Modes, and Latency}
Tool-selection reliability and response latency were quantified separately in~\cite{KnauerIROSA2026} on the same robot and task; we summarize the outcome here and refer to that paper for the protocol and baseline comparisons. Depending on the adaptation type, commands were interpreted as intended in 80--100\% of cases, with all runs completing the task. The dominant failure mode was over-application: the LLM occasionally added an unrequested speed modulation alongside a correct via-point insertion, while still selecting the right tool.

For the system demonstrated at the trade fair (\Cref{tab:parameters}), the average end-to-end adaptation latency was 15.4\,s, roughly half that of a code-generating baseline on the same local backend. Since the demonstration we have reimplemented the pipeline in C++\footnote{\label{fn:pglfd}The C++ implementation will be released as open-source software at \url{https://github.com/DLR-RM/pglfd}.}, which reduces the measured end-to-end latency to 2--3\,s for natural language commands, including LLM inference, and to about 1\,s for physical and graphical adaptations. The bulk of the original latency therefore lay in the surrounding software path rather than in LLM decoding.

Failures are contained by construction, not by model accuracy. Arguments outside the bounds of \Cref{tab:llm_tools} are rejected before the tool executes, and every via-point is inspectable and reversible in the graphical interface (\Cref{fig:via_point_adapt_delete}), irrespective of the modality that created it.
\vspace{0.5em}

\subsection{Web-Based GUI Implementation}
The graphical interface is built with Vue.js~3 for the component framework, Pinia for reactive state management, and Three.js for 3D visualization. The robot digital twin is rendered from URDF models using the \texttt{urdf-loader} library. Communication with the MOMO backend occurs via WebSocket through the LN bridge.

\paragraph{Drag-and-Drop Via-Point Editing.}
When the user enables edit mode, the interface activates 3D picking on the rendered KMP trajectory. Clicking on a trajectory point and dragging it to a new position triggers the following pipeline:
\begin{enumerate}
	\item The frontend sends an ``add via-point'' service call with the point index and new $(x, y, z)$ position via the LN WebSocket bridge,
	\item MOMO inserts the via-point into the KMP model with $\gamma\!=\!10^{-8}$,
	\item the frontend requests the updated model mean and covariance via a second service call,
	\item the trajectory visualization refreshes, showing original (blue) and adapted (yellow) trajectories.
\end{enumerate}
Right-clicking a via-point opens a context menu for adapting (dragging) or deleting it, with the trajectory updating in real time after each modification.

\paragraph{LLM Chat Integration.}
The ChatBox component sends user text to the MOMO backend via the \texttt{set\_llm\_input\_query} service. The frontend subscribes to a notification topic; when the LLM response is ready, it retrieves the answer via \texttt{get\_llm\_answer} and displays it with animated word-by-word rendering. Speech input is handled via a separate speech-to-text service.

\paragraph{Real-Time Digital Twin.}
During trajectory execution, the frontend subscribes to the execution status topic, receiving the current via-point index and execution state. The robot digital twin updates its joint configuration accordingly, with a progress bar indicating execution progress.

\subsection{Integration: Multi-Modal Via-Point Convergence}
A key design insight is that all three modalities converge on the same KMP adaptation mechanism---via-point insertion---through different input paths:
\begin{itemize}
	\item \textbf{Physical interaction}: The energy-tank-based Human Intention Detection (HID) detects intentional corrections per DoF. When the intention index exceeds $h_\text{th}\!=\!0.9$, the corrected pose is automatically inserted as a via-point at the corresponding time step. The KMP's uncertainty modulates how long the user must interact before a correction is registered.
	\item \textbf{Natural language}: The LLM selects the \texttt{AddViaPoints} tool and extracts positions from spatial references in the command (e.g., ``move left at the start'' $\rightarrow$ via-point at $t\!=\!0.1$ with shifted $y$). For obstacle avoidance, the \texttt{AddRepulsion} tool computes repulsion via-points from signed distance fields.
	\item \textbf{Graphical interface}: The user directly places via-points via drag-and-drop on the 3D trajectory visualization.
\end{itemize}
After any via-point insertion (regardless of modality), the KMP recomputes its prediction, and the updated trajectory is immediately available to all other modalities. This enables seamless multi-modal workflows, e.g., adding obstacle avoidance via voice, then fine-tuning the result via drag-and-drop.

\subsection{Human Intention Detection (HID)}
The energy-tank-based intention detection runs as a C++ real-time service; for algorithmic details, see~\cite{Fiorini2026}.
The module requires a torque-controlled robot; in our setup, an external observer estimates contact forces along the entire robot body~\cite{Iskandar2021} rather than relying solely on a wrist-mounted force/torque sensor, enabling more intuitive physical corrections. Sensor signals are filtered at 400\,Hz with dead zones of 7.0\,N (force) and 7.0\,Nm (torque) for the external observer.
The energy tanks are configured per DoF with separate parameters for translational axes (tank size~0.4, energy trigger~0.38, dissipation rate~0.04) and rotational axes (tank size~1.0, energy trigger~0.7, dissipation rate~0.2). When HID is enabled, the robot's collision stop is disabled to permit physical contact.

When the intention index in a specific DoF exceeds the threshold $h_\text{th}$, the detected via-point for that DoF is constructed as follows: for each translational axis where interaction was detected, the current end-effector position (transformed from flange to tool frame) replaces the KMP prediction; axes without detected interaction retain the original KMP-predicted position. If any rotational interaction is detected, the full measured orientation overrides the KMP prediction. The resulting 7-DoF pose (position + quaternion) is paired with the corresponding KMP time index and inserted as a via-point with precision $\gamma\!=\!10^{-8}$.
After trajectory execution, the Cartesian impedance stiffness is reset to $K_f\!=\!1000$\,N/m, $K_t\!=\!100$\,Nm/rad.

\subsection{Virtual Fixtures for Skill Acquisition}
The Virtual Fixtures component~\cite{muehlbauer2024probabilistic,muehlbauer2025unified,muehlbauer2026stabilizing} runs as a Simulink real-time controller parametrized by a Python-based trajectory fixture controller via LN; for algorithmic details, see the cited works. \Cref{fig:fixtures} shows the ellipsoidal fixture visualization used during kinesthetic teaching.

In the experiments, we use GMM-based trajectory fixtures as well as KMP-based velocity fixtures, both learned from demonstrations. Both pipelines share the same preprocessing: pose recordings (position + unit quaternion, stored per time step) are loaded and resampled based on Cartesian distance (threshold~0.001\,m).

For trajectory fixtures, the resampled demonstrations are then aligned via dynamic time warping, and a GMM is fitted to the time-augmented data on a product manifold ($\mathbb{R}^1 \times \mathbb{R}^3 \times \mathcal{S}^3$) with the Riemannian EM algorithm~\cite{zeestraten2017manifold}.
Through GMR, a mean trajectory with associated covariances is extracted, which are inverted to precisions and sent to the real-time controller.
The variable impedance formulation~\cite{muehlbauer2025unified} derives coupled stiffness matrices from these precisions via precision decomposition, providing position-dependent haptic guidance: high-precision regions are stiff to guide the operator, while low-precision regions are compliant to allow deviation.
Passivity of the closed loop is ensured by~\cite{muehlbauer2026stabilizing}.
We use two trajectory fixtures, one for guiding the ring grasping phase and one for the placement on the measurement station.

For velocity fixtures, positional and rotational guidance are learned independently as detailed in~\cite{muehlbauer2025unified}.
Velocities are computed from the resampled poses and encoded in GMMs on $\mathbb{R}^3 \times \mathbb{R}^3$ and $\mathbb{R}^3 \times \mathcal{S}^3 \times \mathbb{R}^3$, which are then used to initialize KMPs with an RBF kernel of length scale~$0.03$.
These velocity fixtures coarsely guide the operator towards the placement location.

Fixture configurations are stored as YAML files containing the GMM parameters (manifold type, means, covariances, priors), kernel configuration, and fixture origin.
The controller supports up to 10 simultaneously active trajectory and velocity fixtures with independent loading, modification, and origin updates.

\subsection{Ergodic Surface Finishing}
The ergodic surface finishing component uses the Spectral Multiscale Coverage (SMC) method for generating coverage trajectories on freeform surfaces, with the ergodic cost evaluated using $K{=}15$ Fourier frequency components per dimension; for algorithmic details (surface parametrization, compliant-disk imprint model, coverage learning from demonstrations), see~\cite{Schneyer2025}.

An impedance controller with force overlay maintains the desired contact normal force.
The LLM tools defined in \Cref{tab:llm_tools} set runtime properties on the ergodic controller via a \textit{links and nodes} service call: \texttt{SetVelocity} sets the velocity limit (range $[3, 16]$, default~6), \texttt{SetForce} sets the contact normal force setpoint (range $[5, 30]$\,N, default~15), \texttt{SetStiffness} sets the translational surface-tangential stiffness components (range $[500, 2000]$\,N/m, default~1000, while the normal component remains fixed at 800\,N/m), and \texttt{SetExecState} switches the state machine between idle and free-floating mode to pause or resume execution.
Since trajectories are generated online, the operator can physically interact during execution: end-effector stiffness is varied with the distance to the surface, allowing the operator to lift the tool and reposition the robot to a new location on the workpiece, after which execution resumes while preserving the full coverage history.

\subsection{Configuration Parameters}
\Cref{tab:parameters} lists the key configuration parameters.
\begin{table}[H]
	\centering
	\caption{Key configuration parameters of the framework components.}
	\label{tab:parameters}
	\footnotesize
	\setlength{\tabcolsep}{3pt}
	\begin{tabular}{@{}lll@{}}
		\toprule
		\textbf{Parameter} & \textbf{Value} & \textbf{Component} \\
		\midrule
		\rowcolor{Light1}
		GMM components & 12 & KMP \\
		Sample points $N$ & 500 & KMP \\
		\rowcolor{Light1}
		Kernel (Mat\'ern $\nu\!=\!5/2$), $l$ & 0.1 & KMP \\
		Regularization $\lambda_1$, $\lambda_2$ & 0.1, 1 & KMP \\
		\rowcolor{Light1}
		Via-point precision $\gamma$ & $10^{-8}$ & KMP \\
		\midrule
		Intention threshold $h_\text{th}$ & 0.9 & HID \\
		\rowcolor{Light1}
		Tank size $E_{\max}$ (trans.\ / rot.) & 0.4 / 1.0 & HID \\
		Energy trigger $E^*$ (trans.\ / rot.) & 0.38 / 0.7 & HID \\
		\rowcolor{Light1}
		Dissipation rate $P_d$ (trans.\ / rot.) & 0.04 / 0.2 & HID \\
		Reset stiffness $K_f$ / $K_t$ & 1000 / 100 & HID \\
		\midrule
		\rowcolor{Light1}
		Resampling threshold & 0.001\,m & VF \\
		RBF kernel length scale $l_v$ & 0.03 & VF \\
		\rowcolor{Light1}
		Max.\ active fixtures & 10 & VF \\
		\midrule
		\rowcolor{Light1}
		LLM model & Qwen2.5-VL-72B-Instruct & LLM \\
		LLM deployment & Local (GPU) & LLM \\
		\rowcolor{Light1}
		Avg.\ response time & 15.4\,s & LLM \\
		\bottomrule
	\end{tabular}
\end{table}

\subsection{Open-Source Resources and Reproducibility}
\begin{table}[H]
	\centering
	\caption{Open-source resources accompanying/used in this paper.}
	\label{tab:opensource}
	\footnotesize
	\setlength{\tabcolsep}{2pt}
	\begin{tabular}{@{}lll@{}}
		\toprule
		\textbf{Resource} & \textbf{License} & \textbf{Content} \\
		\midrule
		\rowcolor{Light1}
		\systemname{}\textsuperscript{\ref{fn:irosa}} & MIT & LLM tool arch.\ + examples \\
		KMP/TP-KMP\textsuperscript{\ref{fn:kmp}} & MIT & Motion primitives \\
		\rowcolor{Light1}
		\textit{links and nodes} (LN)\textsuperscript{\ref{fn:ln}} & GPLv3 & Middleware \\
		PGLfD\textsuperscript{\ref{fn:pglfd}} & MIT & C++ reimpl.\ (planned) \\
		\bottomrule
	\end{tabular}
\end{table}
\Cref{tab:opensource} lists all open-source resources used with this work.

The \textbf{\systemname{}}\footnote{\label{fn:irosa}\url{https://github.com/DLR-RM/IROSA}} provides the tool-based LLM architecture including tool definitions, parameter validation, KMP model, and integration with OpenAI-compatible function calling APIs. It includes three example scripts demonstrating speed adaptation, via-point insertion, and obstacle avoidance, each runnable without hardware dependencies. The open-source release replaces the proprietary LLM backend with a standard interface for reproducibility.

\textbf{KMP/TP-KMP}\footnote{\label{fn:kmp}\url{https://github.com/DLR-RM/interactive-incremental-learning}} provides the motion primitive implementation with interactive incremental learning, four reproducible experiments, and demonstration data.

\textbf{Links and nodes}~\cite{LinksAndNodes2020} is the open-source middleware connecting all components, providing process management, pub/sub topics, and request/response services.

\end{multicols}

\clearpage
\begin{figure}[ht]
	\centering
	\includegraphics[width=\overviewfigwidth]{hfi_overview.png}
	\caption{Overview of the Human Factory Interface consisting of four views: (a)~Factory Overview, (b)~Task Container Design View, (c)~Workcell Viewer, and (d)~Ontology Viewer.}
	\label{fig:hfi}
\end{figure}

\section{Human Factory Interface}
\label{sec:hfi}

\begin{multicols}{2}
The Human Factory Interface (HFI) is a web-based graphical interface (implementation details in Sec.~\ref{sec:implementation}) that enables operators to monitor, configure, and control robotic workcells, i.e., workspaces comprising one or more robots together with the tools, fixtures, and objects they operate on. In our setup, the HFI manages a \emph{variable workcell}~\cite{Bachmann2023}, in which fixtures and assets can be reconfigured for different tasks and the layout is registered in the knowledge graph so that the digital twin always reflects the current physical arrangement. The system is backed by a knowledge graph~\cite{schafer2021flexible} that stores the relationships between workcells, robots, physical assets, and a library of reusable robot skills---parameterizable building blocks such as \textit{pick}, \textit{place}, or \textit{insert} that encapsulate individual robot actions~\cite{eiband2024unifying}. Non-expert users can compose robot programs by sequencing and parameterizing these skills, either through a graphical interface or by recording demonstrations~\cite{eiband2023collaborative,eiband2024unifying}. As shown in Fig.~1 in the main paper, the HFI observes and receives notifications from the knowledge graph, and communicates directly with the robot for execution and monitoring.

The HFI is organized into four views, as shown in \Cref{fig:hfi}:

\paragraph{Factory Overview.}
The factory overview (\Cref{fig:hfi}a) displays all registered workcells and their operational states. Operators can monitor which skill is currently executing, pause or stop execution, and investigate warnings or failures.

\paragraph{Task Container Design View.}
The task container design view (\Cref{fig:hfi}b) allows users to compose task sequences by selecting skills from the library, setting their parameters (e.g., target positions, velocities), and arranging them into executable programs.

\paragraph{Workcell Viewer.}
The workcell viewer (\Cref{fig:hfi}c) is the primary view used in this work. It provides a 3D visualization of the robotic workcell, rendering the robot and all registered assets as a digital twin. This view supports direct manipulation and has been extended with the skill visualization and interaction capabilities described in the following section.

\paragraph{Ontology Viewer.}
The ontology viewer (\Cref{fig:hfi}d) visualizes the knowledge graph as an interactive node-edge diagram, showing available resources, active configurations, and the relationships between system components~\cite{schafer2021flexible}.
\end{multicols}

\clearpage
\section{MOMO Interface Screenshots}
\label{sec:momo_screenshots}

\begin{figure}[ht]
	\centering
	\includegraphics[width=\screenshotfigwidth]{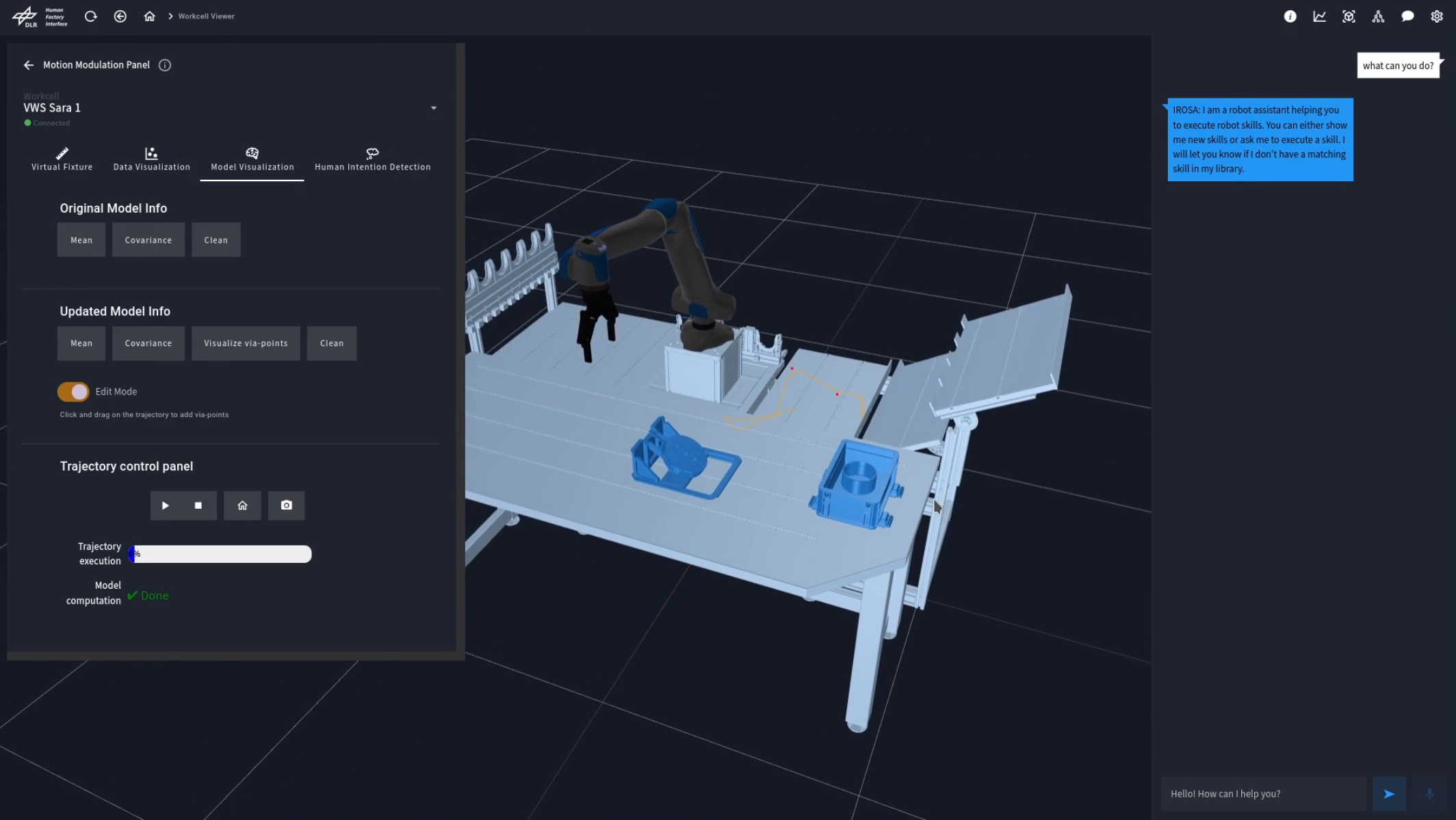}
	\caption{Overview of the MOMO web-based interface. The \systemname{} is integrated as a chat window (right), enabling natural language skill adaptation through text and speech. The left panel shows the Model Visualization tab with controls for the original and updated model, edit mode, and trajectory execution. The 3D viewer displays the robot cell with the digital twin.}
	\label{fig:momo_irosa_chat}
\end{figure}

\begin{figure}[ht]
	\centering
	\includegraphics[width=\screenshotfigwidth]{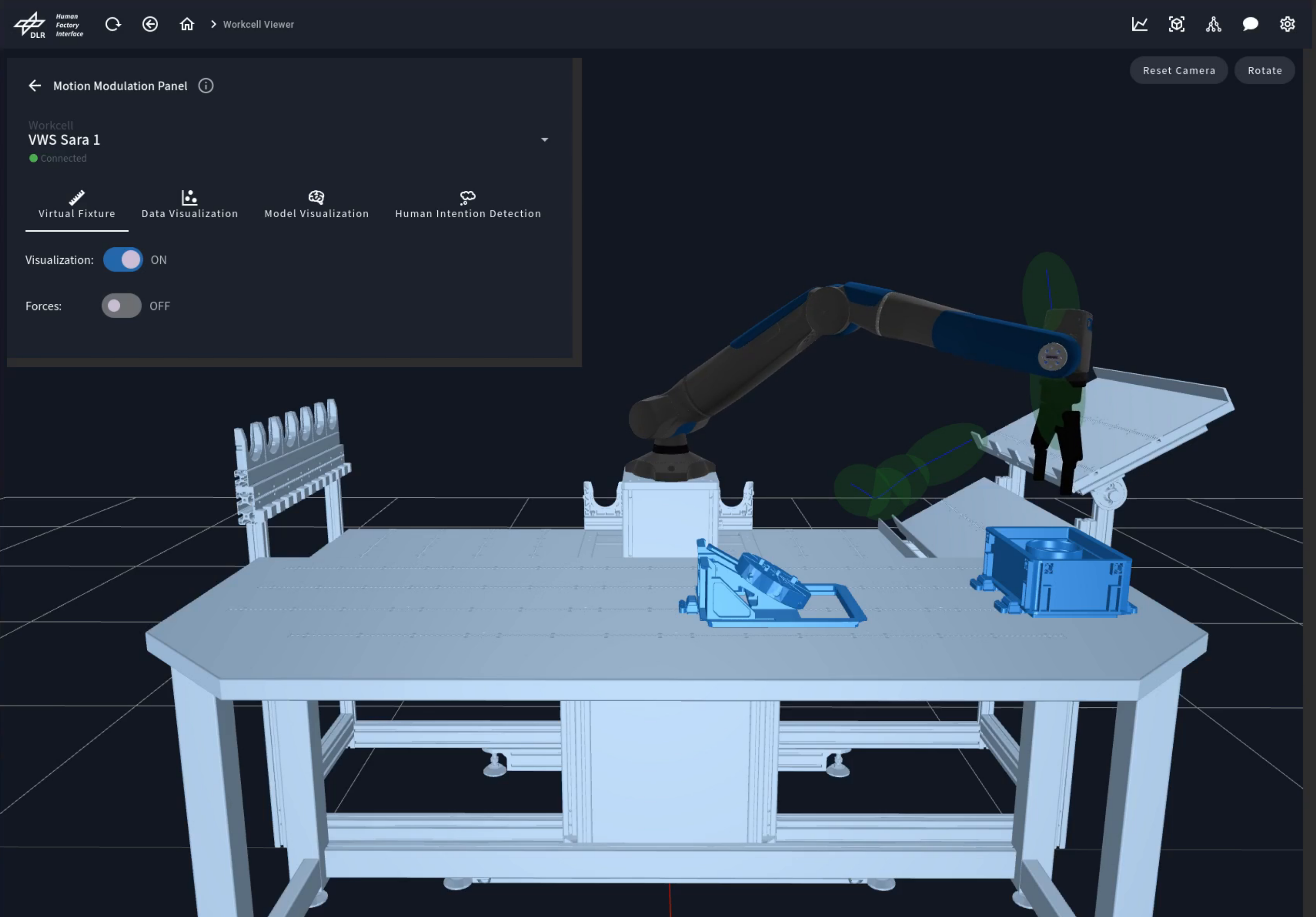}
	\caption{Default Fixtures tab showing demonstration recording with Virtual Fixtures. The green, semi-transparent ellipsoids overlaid on the trajectory represent the probabilistic Virtual Fixtures that provide haptic guidance to the operator during kinesthetic teaching.}
	\label{fig:fixtures}
\end{figure}

\begin{figure}[ht]
	\centering
	\includegraphics[width=\screenshotfigwidth]{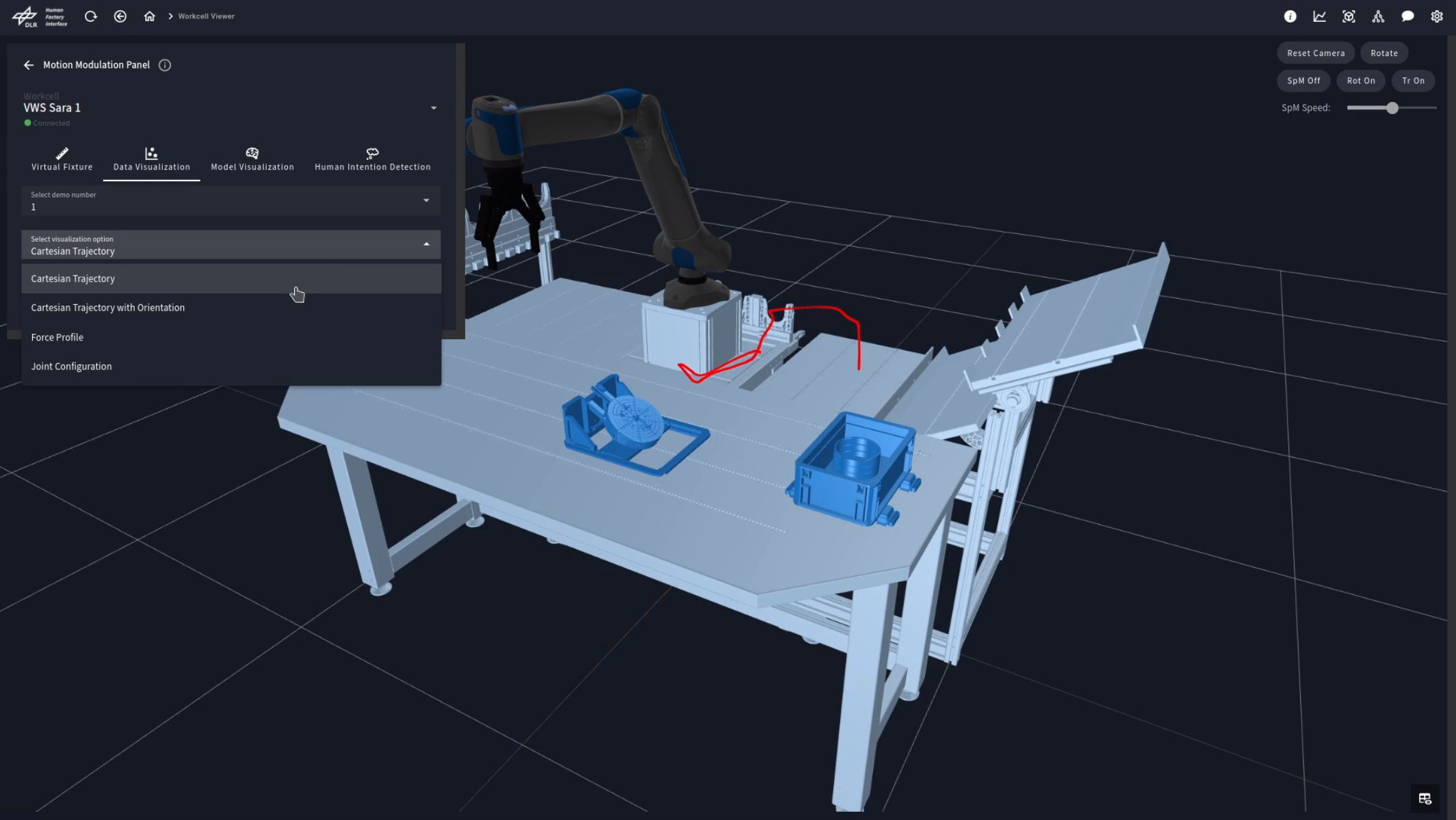}
	\caption{Data Visualization tab of the web-based interface. A recorded demonstration is selected and its Cartesian trajectory is visualized in red (flange position) projected into the 3D robot cell. The dropdown menu provides additional visualization options: Cartesian Trajectory with Orientation, Force Profile, and Joint Configuration.}
	\label{fig:momo_data_visualization}
\end{figure}

\begin{figure}[ht]
	\centering
	\includegraphics[width=\screenshotfigwidth]{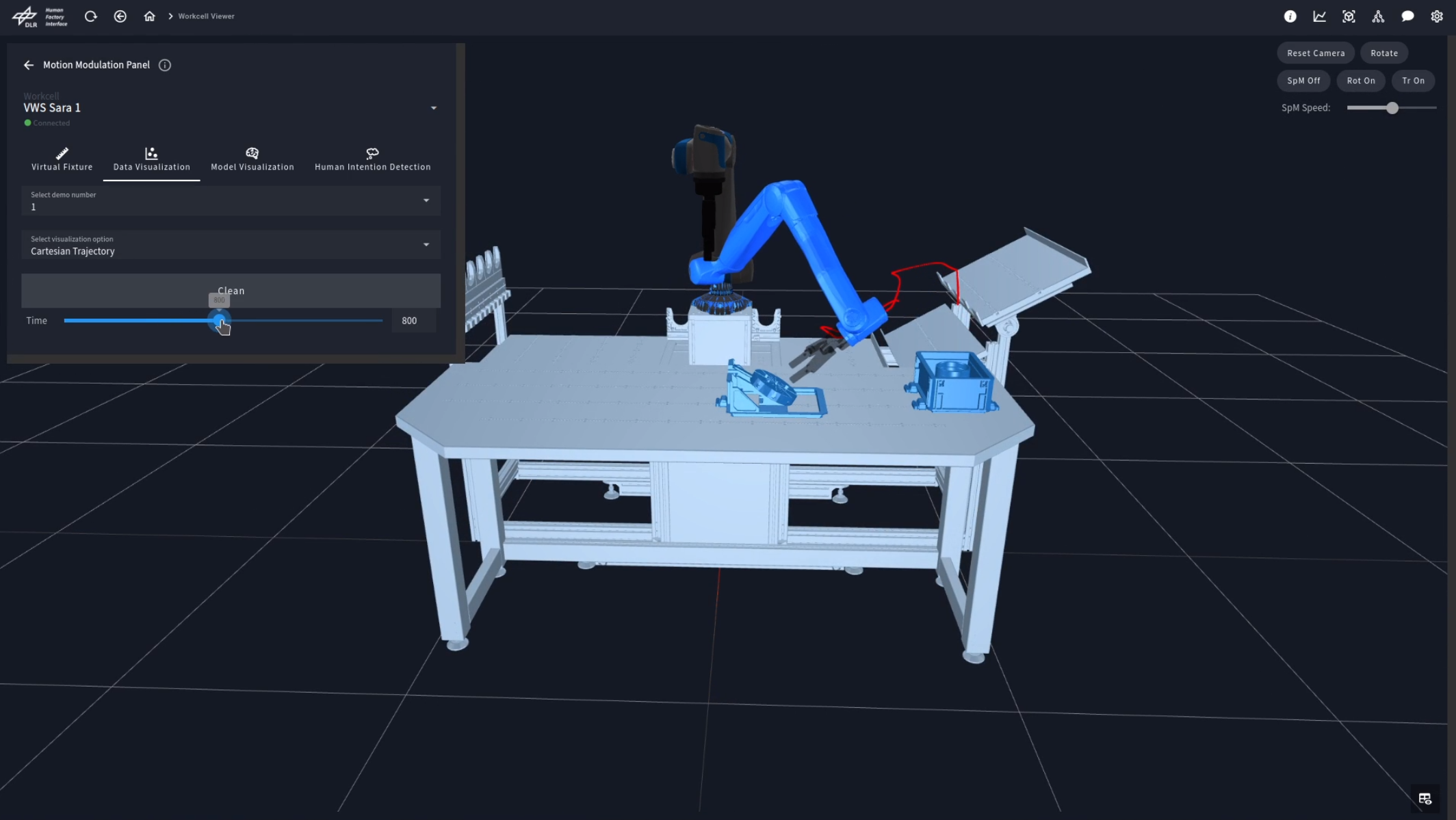}
	\caption{Data Visualization tab with the Joint Configuration option selected. A blue robot hologram shows the robot pose at the currently selected time step of the demonstration.}
	\label{fig:momo_data_visualisation_joint}
\end{figure}

\begin{figure}[ht]
	\centering
	\includegraphics[width=\screenshotfigwidth]{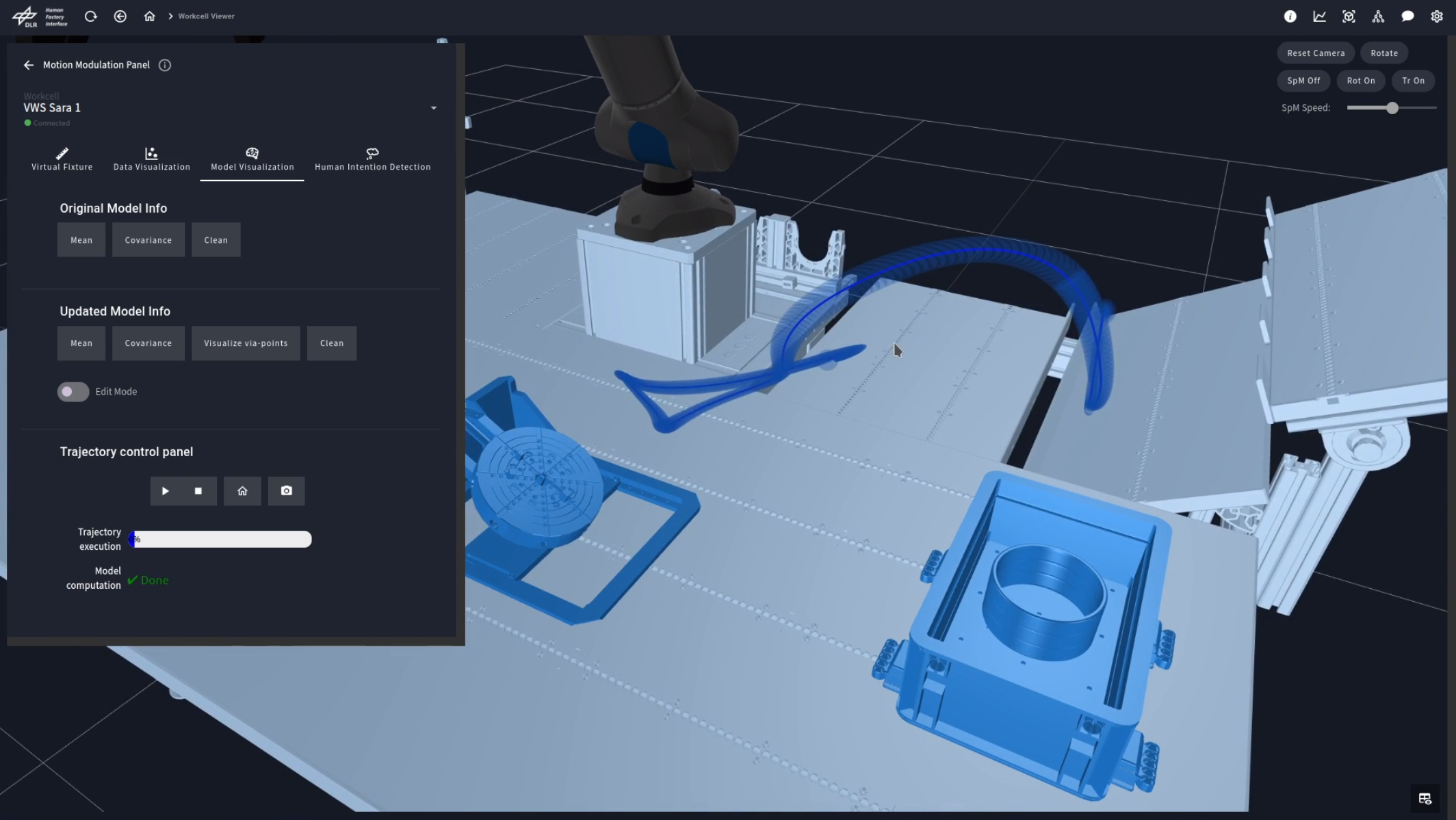}
	\caption{Model Visualization tab showing the original KMP mean and covariance in blue projected into the robot cell. The panel provides controls for visualizing both the original and updated model (e.g., after via-point insertion), an edit mode toggle for interactive trajectory editing, and a trajectory control panel for starting execution and monitoring progress.}
	\label{fig:momo_model_visualization}
\end{figure}

\begin{figure}[ht]
	\centering
	\includegraphics[width=\screenshotfigwidth]{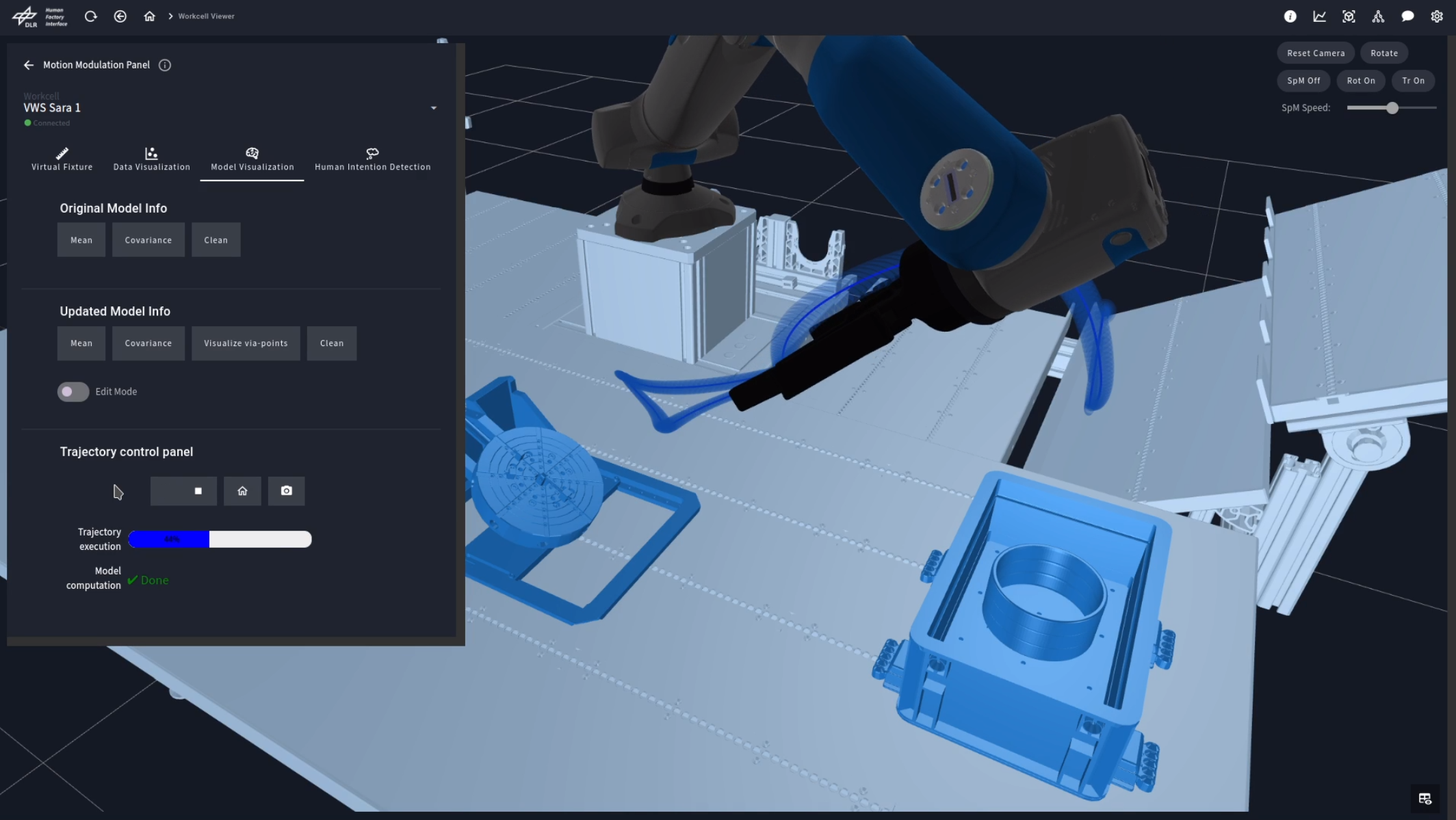}
	\caption{Model Visualization tab during trajectory execution. The trajectory control panel shows the execution progress, and the digital twin of the robot moves in sync with the real robot.}
	\label{fig:momo_model_vis_exec}
\end{figure}

\begin{figure}[ht]
	\centering
	\includegraphics[width=\screenshotfigwidth]{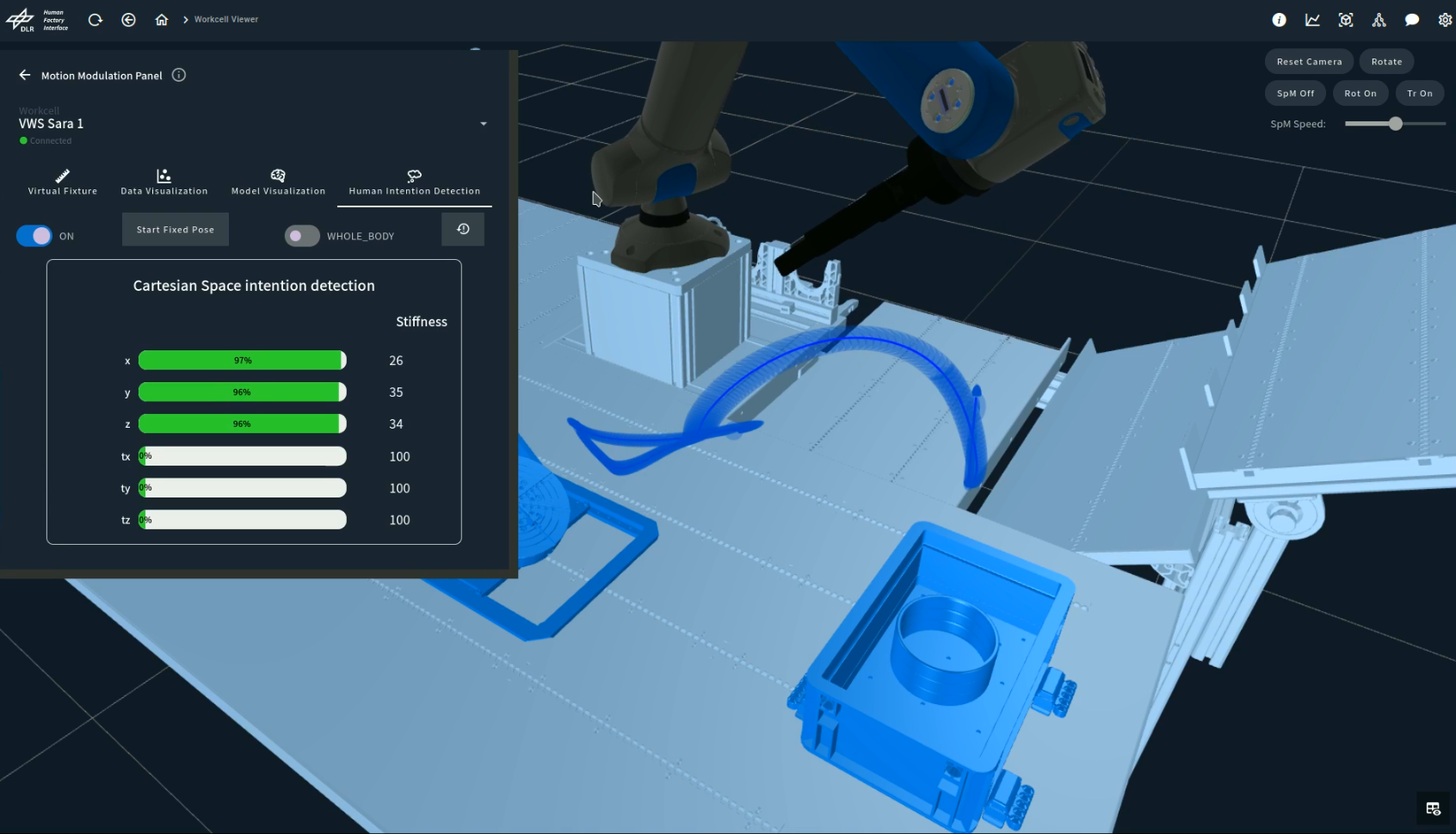}
	\caption{\textit{Human Intention Detection} tab. When enabled, the interface displays for each Cartesian DoF ($x$, $y$, $z$, $\theta_x$, $\theta_y$, $\theta_z$) the intention index, proportional to the energy tank value, along with the corresponding robot stiffness, which is inversely proportional to the intention index, allowing the operator to monitor physical human-robot interaction in real time.}
	\label{fig:momo_human_intention_detection}
\end{figure}

\begin{figure}[ht]
	\centering
	\includegraphics[width=\screenshotfigwidth]{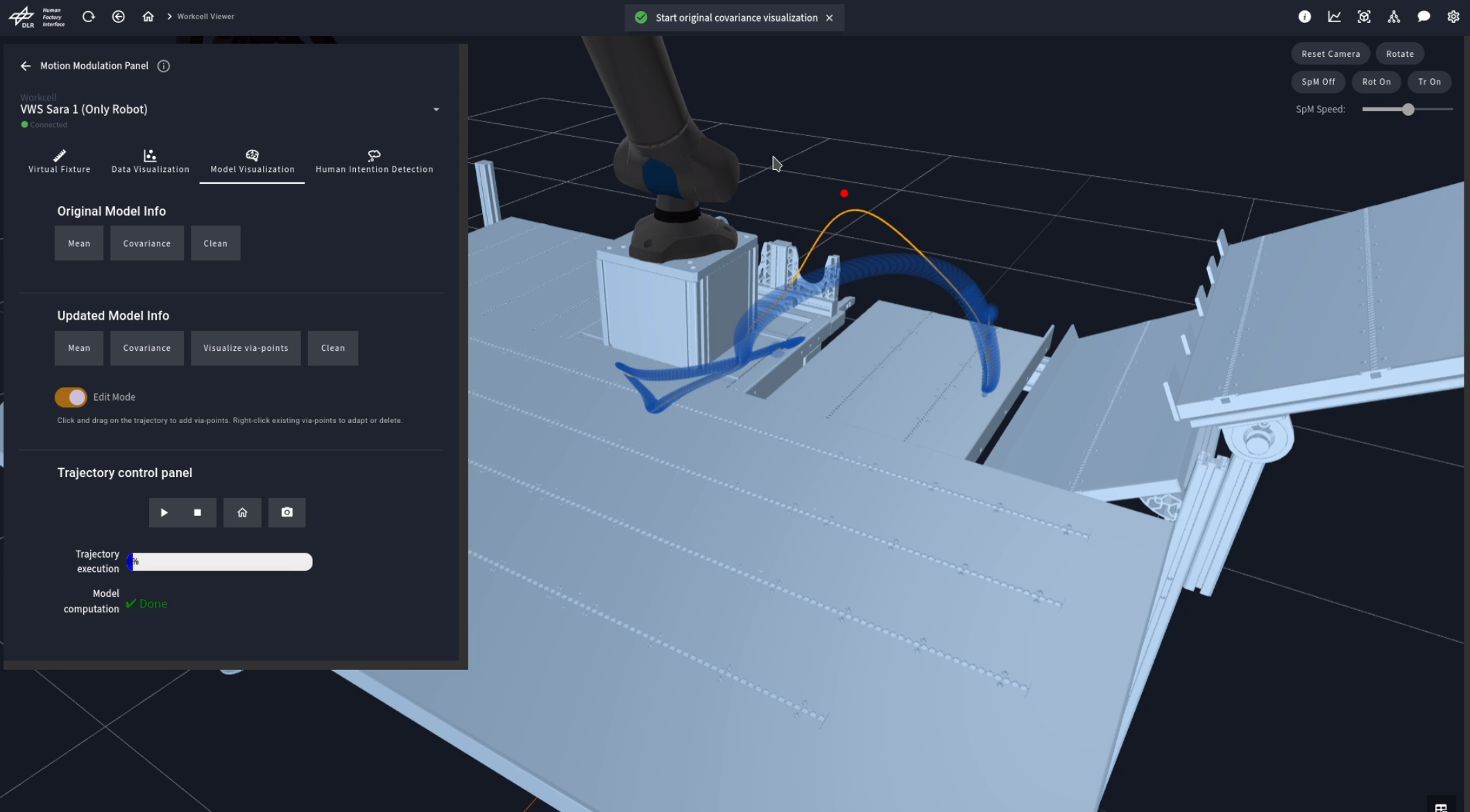}
	\caption{Model Visualization showing a via-point adaptation. The original KMP mean and covariance are displayed in blue, the adapted mean trajectory in yellow, and the set via-point in red.}
	\label{fig:via_point_set}
\end{figure}

\begin{figure}[ht]
	\centering
	\includegraphics[width=\screenshotfigwidth]{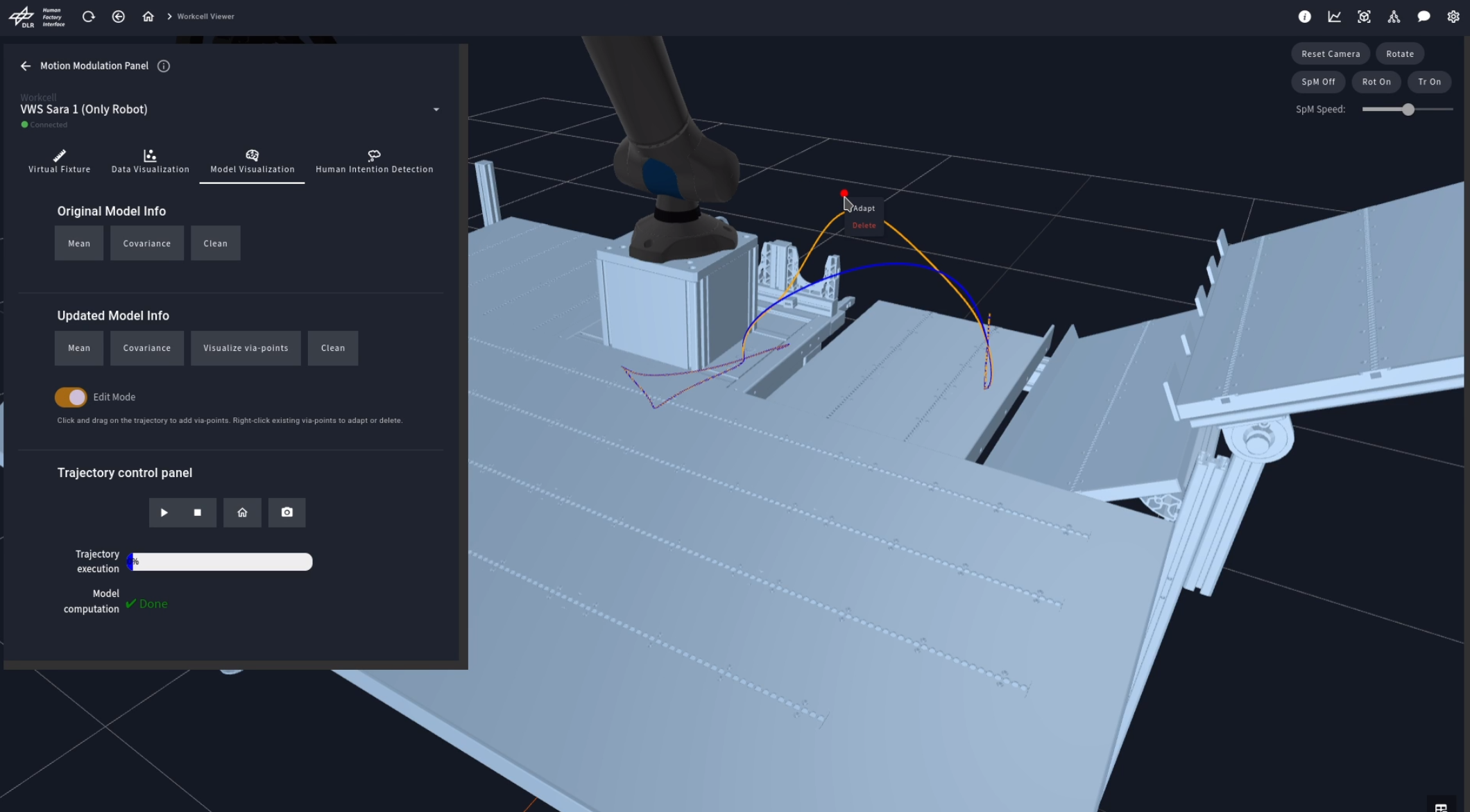}
	\caption{Context menu for via-point manipulation. Right-clicking on a via-point opens a menu with options to adapt (move with the mouse) or delete the via-point, with the trajectory updating accordingly in real time.}
	\label{fig:via_point_adapt_delete}
\end{figure}

\begin{figure}[ht]
	\centering
	\includegraphics[width=\screenshotfigwidth]{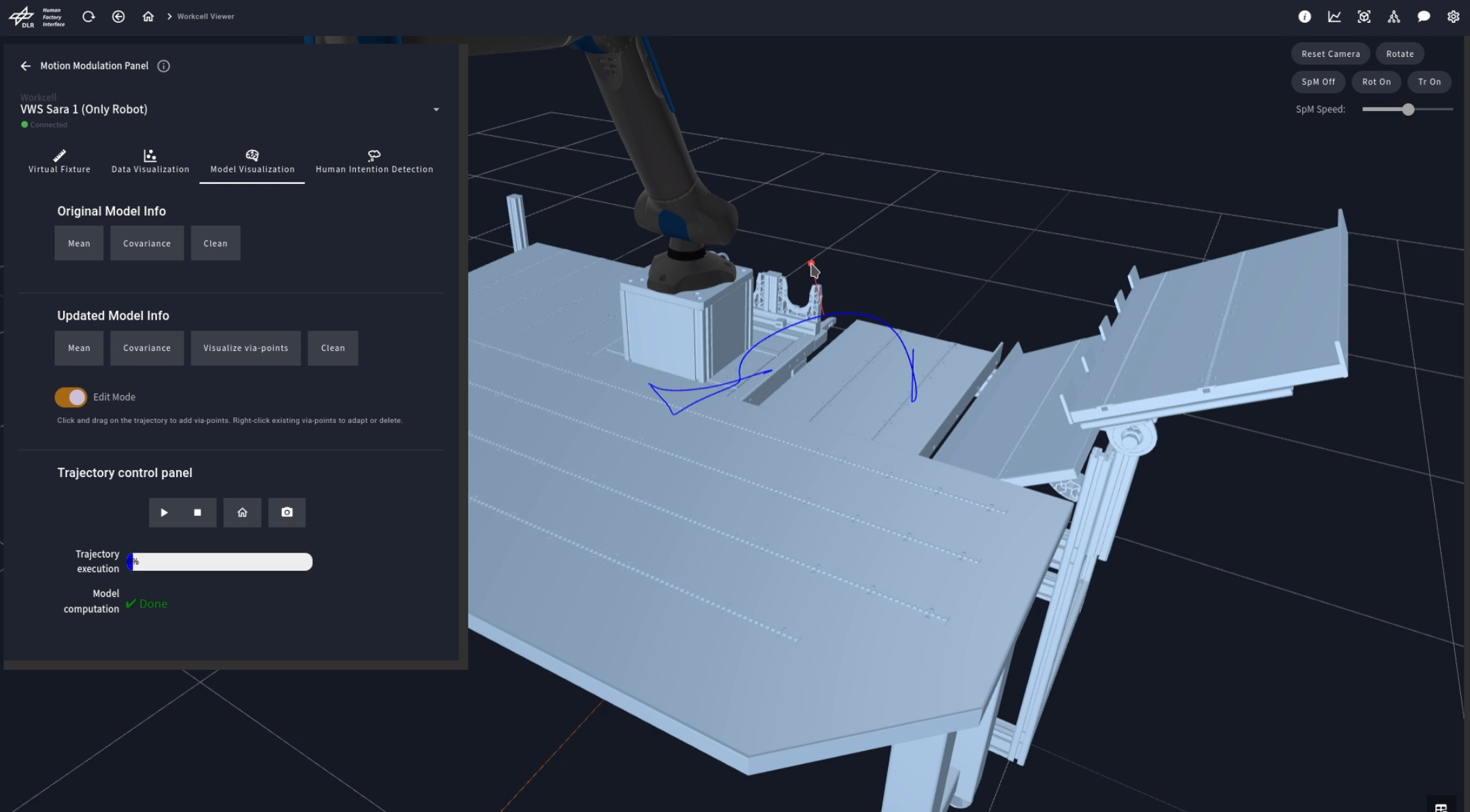}
	\caption{Drag-and-drop via-point creation. The user clicks on a trajectory point and drags it to a new position, creating a via-point directly through the graphical interface.}
	\label{fig:drag_drop_hfi}
\end{figure}

\begin{figure}[ht]
	\centering
	\includegraphics[width=\screenshotfigwidth]{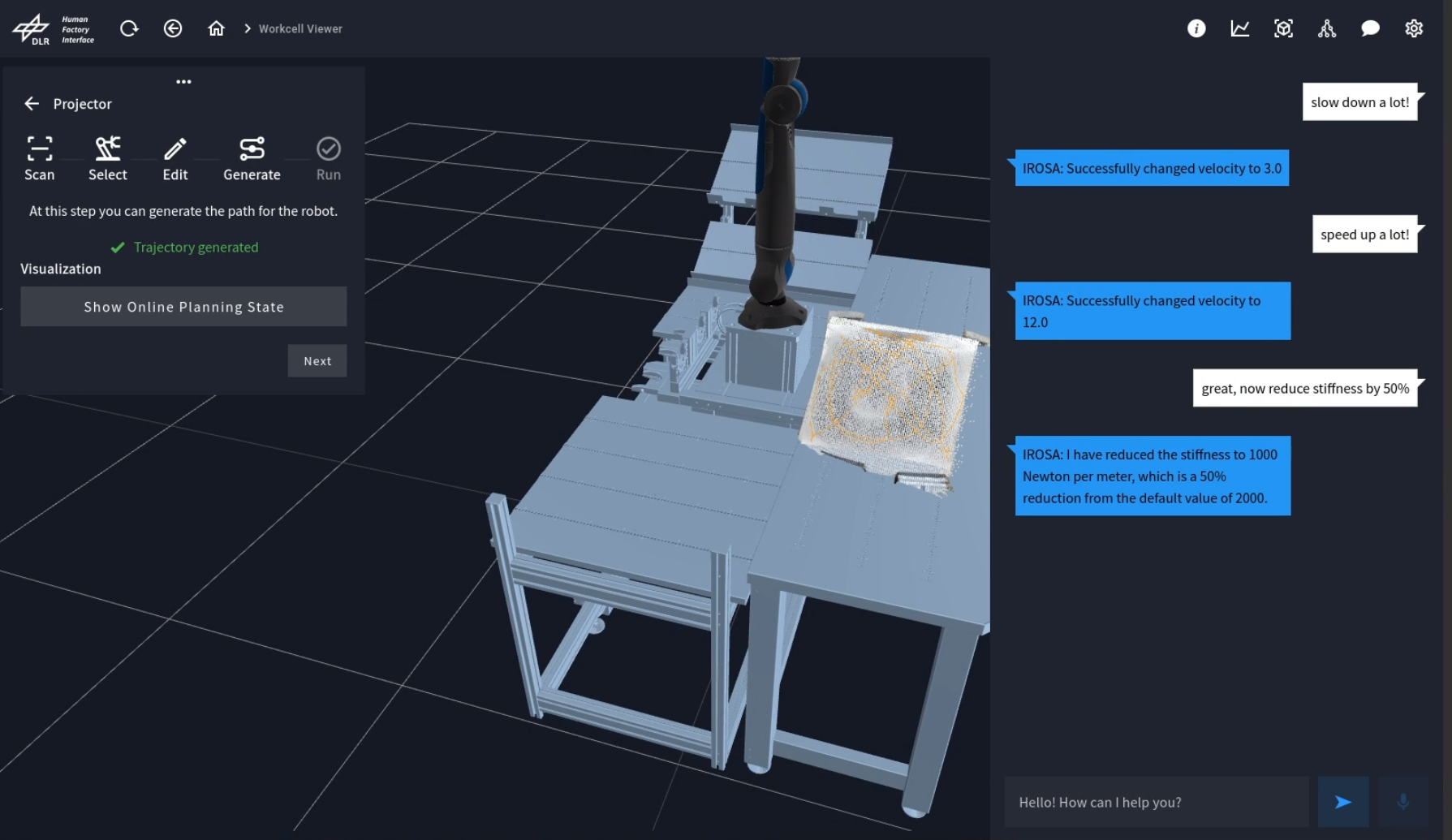}
	\caption{Ergodic surface finishing skill with integrated \systemname{} chat interface. This demonstrates that the framework, including the LLM-based natural language adaptation, generalizes beyond KMP-based approaches to other skill representations such as ergodic control. The chat shows example commands for velocity and stiffness adaptation.}
	\label{fig:ergodic_llm}
\end{figure}

\begin{multicols}{2}
The Motion Modulation Panel, i.e., the graphical interface of the MOMO module (Sec.~\ref{sec:implementation}), extends the HFI Workcell Viewer (Sec.~\ref{sec:hfi}) with skill visualization and interactive adaptation. The following screenshots (\Cref{fig:momo_irosa_chat}--\Cref{fig:ergodic_llm}) illustrate its features.
\end{multicols}

\end{document}